\documentclass{article}

    \usepackage[final]{neurips_2025}

\usepackage[utf8]{inputenc} % allow utf-8 input
\usepackage[T1]{fontenc}    % use 8-bit T1 fonts
\usepackage{hyperref}       % hyperlinks
\usepackage{url}            % simple URL typesetting
\usepackage{booktabs}       % professional-quality tables
\usepackage{amsfonts}       % blackboard math symbols
\usepackage{nicefrac}       % compact symbols for 1/2, etc.
\usepackage{microtype}      % microtypography
\usepackage{xcolor}         % colors

\usepackage{amsmath}
\usepackage{multirow}
\usepackage{makecell}
\usepackage{adjustbox}
\usepackage{graphicx}
\usepackage{subfigure}
\usepackage[table]{xcolor}
\usepackage{colortbl}
\usepackage{comment} 
\usepackage{pifont}
\usepackage{enumitem}
\title{Mitigating Hallucination Through Theory-Consistent Symmetric Multimodal Preference Optimization}

\author{%
\textbf{Wenqi Liu}$^1$ \quad
\textbf{Xuemeng Song}$^2$\footnotemark[2] \quad
\textbf{Jiaxi Li}$^3$ \quad
\textbf{Yinwei Wei}$^1$ \\
\textbf{Na Zheng}$^4$ \quad
\textbf{Jianhua Yin}$^1$\footnotemark[2] \quad 
\textbf{Liqiang Nie}$^5$ \\
$^1$Shandong University \ 
$^2$Southern University of Science and Technology \ 
$^3$University of Georgia \\
$^4$National University of Singapore \ 
$^5$Harbin Institute of Technology (Shenzhen) \\
\texttt{liuwq\_bit@outlook.com, sxmustc@gmail.com, jhyin@sdu.edu.cn} \\
}

\begin{document}
\renewcommand{\thefootnote}{\fnsymbol{footnote}}
\footnotetext[2]{Corresponding authors.}

\maketitle

\begin{abstract}
Direct Preference Optimization (DPO) has emerged as an effective approach for mitigating hallucination in Multimodal Large Language Models (MLLMs). Although existing methods have achieved significant progress by utilizing vision-oriented contrastive objectives for enhancing MLLMs' attention to visual inputs and hence reducing hallucination, they suffer from non-rigorous optimization objective function and indirect preference supervision. To address these limitations, we propose a \textbf{Sym}metric \textbf{M}ultimodal \textbf{P}reference \textbf{O}ptimization (SymMPO), which conducts symmetric preference learning with direct preference supervision (i.e., response pairs) for visual understanding enhancement, while maintaining rigorous theoretical alignment with standard DPO. In  addition to conventional ordinal preference learning, SymMPO introduces a preference margin consistency loss to quantitatively regulate the preference gap between symmetric preference pairs. Comprehensive evaluation across five benchmarks demonstrate SymMPO's superior performance, validating its effectiveness in hallucination mitigation of MLLMs. Our codes are available at~\href{https://github.com/Liuwq-bit/SymMPO}{https://github.com/Liuwq-bit/SymMPO}.
\end{abstract}

% Recently, Large Language Models (LLMs) have achieved remarkable success in natural language understanding and generation tasks~\cite{achiam2023gpt,touvron2023llama}. By incorporating visual encoders and cross-modal alignment modules, Multimodal Large Language Models (MLLMs)~\cite{liu2023visual, openai2023gpt4v, bai2023qwen} have extended the capabilities of LLMs to the multimodal domain, enabling applications such as visual question answering (VQA), image captioning, and multimodal dialogue systems. Despite their impressive performance, MLLMs often face significant challenges with hallucination~\cite{bai2024hallucination, gunjal2024detecting, guan2024hallusionbench}, where they produce outputs that are linguistically fluent but factually inaccurate—especially in tasks requiring precise interpretation of visual inputs.
\section{Introduction}\label{section:introduction}
\begin{figure}
    \centering
    \includegraphics[width=1.0\linewidth]{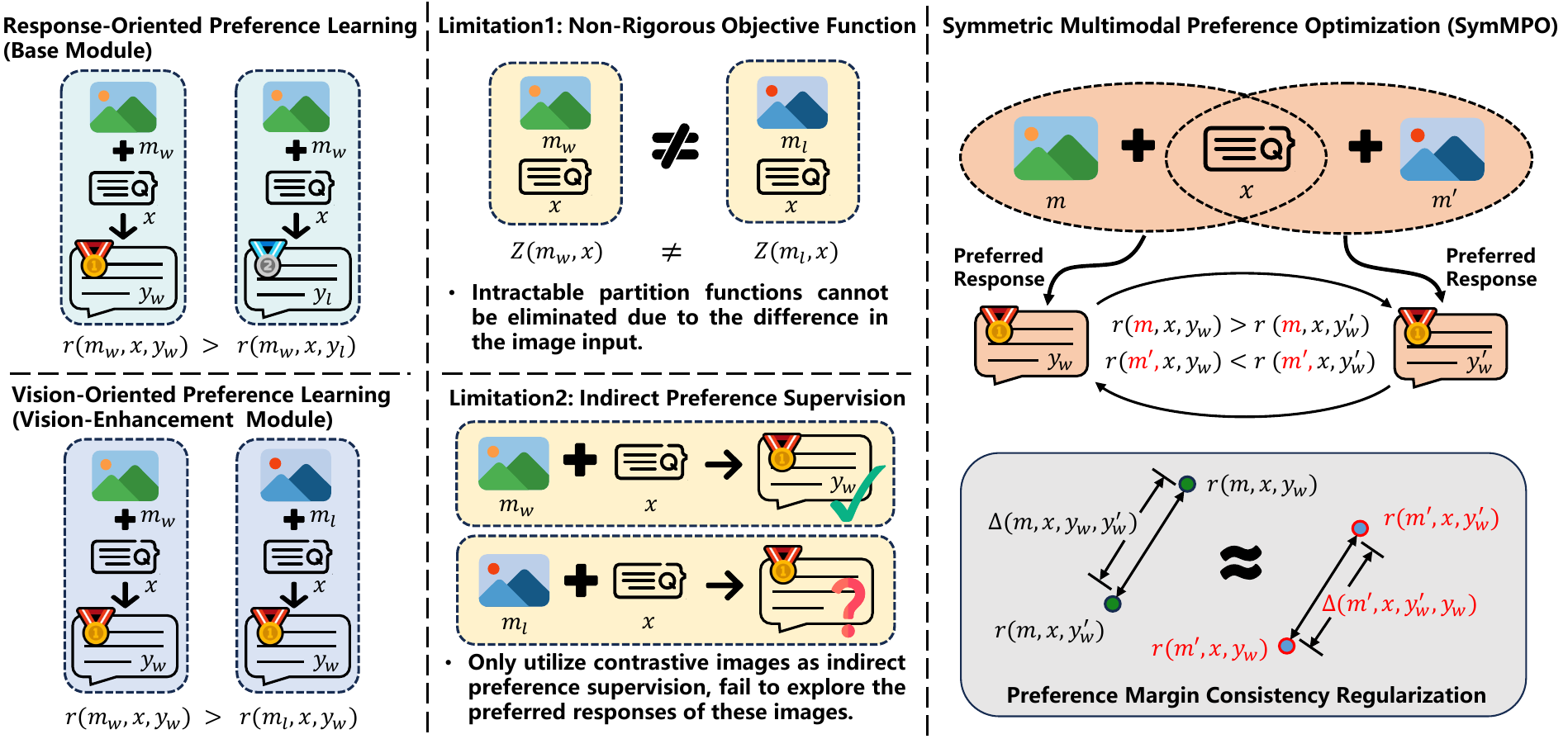}
    \caption{Existing methods focus on response- and vision-oriented preference learning (left), where the former serves as the base module, while the latter is optional, specifically for vision understanding enhancement. $r(m,x,y)$ represents the reward for generating response $y$ given the input $(m, x)$.
    However, current methods face two key limitations: non-rigorous objective function and indirect preference supervision (middle). To address these, we propose SymMPO, which leverages symmetric pairwise preference optimization using contrastive images and their preferred responses, alongside preference margin consistency regularization designed specifically for symmetric optimization (right).}
    \label{fig:symmpo}
    \vspace{-1em}
\end{figure}

Recently, Large Language Models (LLMs) have demonstrated remarkable success in natural language understanding and generation tasks~\cite{achiam2023gpt,touvron2023llama}. By integrating visual encoders and cross-modal alignment modules to LLMs, Multimodal Large Language Models (MLLMs)~\cite{liu2023visual, openai2023gpt4v, bai2023qwen} have extended capabilities of LLMs to the multimodal field, enabling applications such as visual question answering (VQA), image captioning, and multimodal dialogue systems. Despite their impressive performance, MLLMs often suffer from hallucination problems~\cite{gunjal2024detecting, guan2024hallusionbench}, generating outputs inconsistent with the given image input or not relevant to the input textual prompt~\cite{bai2024hallucination}. 
Due to its impressive performance in improving response quality by aligning LLMs with human preferences, i.e., guiding LLMs to generate outputs humans would judge as better (e.g., more safe or contextually appropriate), Direct Preference Optimization (DPO)~\cite{rafailov2023direct} has been adapted to address hallucination issues in MLLMs~\cite{yu2024rlhf, yu2024rlaif, zhang2024automated, zhao2023beyond}.  %Specifically, Originally designed to align LLMs with human preferences, DPO ensures that LLMs generate outputs humans would judge as better (e.g., more safe, or contextually appropriate). 
%The core of DPO lies in the construction of preference response pairs in the form of ($y_m$, $y_l$) for the given textual input $x$, and the optimization objective that maximizes the likelihood of ranking $y_w$ higher than $y_l$ given $x$. 
%To adapt DPO to mitigating hallucinations, 
Existing methods~\cite{yu2024rlhf, sun2023aligning} first construct a preference pair consisting of a hallucination-free response $y_w$ and a hallucinated response $y_l$ for a given multimodal input, including an image $m$ and a textual prompt $x$ that specifies the generation task (e.g., VQA), and then perform DPO-based response-oriented preference learning (see Figure~\ref{fig:symmpo}).% using multimodal inputs, i.e., an image $m$ and a textual prompt $x$ that specifies the generation task (e.g., VQA). 
To strengthen MLLMs' attention to visual inputs, recent studies further incorporate a DPO-based vision-oriented preference learning component~\cite{wang2024mdpo, xie2024v, jiang2024modality, fu2025chip, yang2025mitigating}, as shown in Figure~\ref{fig:symmpo}. This component leverages contrastive triplet pairs $(m_w, x, y_w)$ and $(m_l, x, y_w)$ with only images varied, and aims to make the likelihood of generating $y_w$ given $(m_w, x)$ higher than that given ($m_l$, x). Here, $m_l$ is a modified version of $m_w$, obtained by either applying transformations to $m_w$ or adding noise to it. %This strategy encourages the model to distinguish subtle visual differences between $m_w$ and $m_l$, ultimately leading to more accurate and visually-grounded responses. 

While current methods have demonstrated promising results in MLLM hallucination mitigation by incorporating vision-oriented preference learning, they suffer from two key limitations.

\textbf{Limitation 1. Non-Rigorous Objective Function.}
For deriving the objective function, existing works directly replace the shared image input $m$ in the pairwise likelihoods of the response-oriented preference learning objective function with its counterparts (i.e., $m_w$ and $m_l$) without rigorous mathematical justification.
However, through a careful analysis of the standard multimodal DPO loss derivation, we find that the partition functions $Z(m_w,x)$ and $Z(m_l,x)$---which involve summing over all possible responses and are computationally expensive to estimate in the implicit reward formulation---cannot be directly canceled out in the loss function when the image inputs differ (i.e., $m_w\neq m_l$). In response-oriented preference learning, these partition functions naturally vanish due to their shared input $m$. However, this cancellation does not hold for vision-oriented preference learning, where the contrastive image inputs $m_w$ and $m_l$ are inherently distinct.  
Nevertheless, existing methods directly assume their eliminability, which results in a misaligned objective function that deviates from the theoretical formulation, ultimately limiting model performance~\cite{du2020improved}. 
A rigorous mathematical analysis regarding partition functions is provided in Appendix~\ref{appendix:discussion_of_partition_function}.

%2) In addition to this methodological gap, a critical theoretical issue arises in the vision-oriented contrastive mechanism. In the derivation of standard DPO~\cite{rafailov2023direct}, the Bradley-Terry model enables the elimination of the intractable partition functions in the implicit reward calculation, as these partition functions share identical contextual inputs for both preferred and less preferred responses. However, when adapting DPO to vision-oriented contrastive learning, the assumption of partition function elimination becomes invalid due to the vary visual content between the original and contrastive images. This discrepancy introduces a fundamental inconsistency between theoretical formulation and practical implementation. As a result, the optimization objective derived from the current vision-oriented contrastive mechanism may fail to align with its theoretical foundation, potentially compromising the effectiveness of the entire framework.

\textbf{Limitation 2. Indirect Preference Supervision.}
As shown in Figure~\ref{fig:symmpo}, existing methods conduct DPO with two image-text-response triplets with fixed prompt and response but contrastive images for vision-oriented preference learning. This design fundamentally relies on contrastive images for preference alignment, which misaligns with DPO's core principle of reward formulation via paired responses (i.e., preferred vs. less preferred answers). This misalignment fails to explicitly model the direct preference relationship between accurate and hallucinated responses conditioned on the given image, leading to suboptimal visual understanding capabilities. In fact, we can additionally construct a preferred response for the contrastive image $m_l$ using the same prompt for $m_w$. Since $m_w$ and $m_l$ share high similarity, their corresponding preferred responses naturally exhibit strong semantic alignment with subtle variations---making them ideal hard negatives for each other. Such contrastive response pairs with minor differences compel the model to perform more thorough multimodal interpretation, thereby effectively reducing hallucinations.

To address these limitations, we propose \textbf{Sym}metric \textbf{M}ultimodal \textbf{P}reference \textbf{O}ptimization (SymMPO), to reduce hallucinations in MLLMs. Unlike existing methods that solely rely on contrastive images to conduct DPO-based vision-oriented preference alignment, SymMPO additionally introduces the preferred response of the contrastive image given the same prompt of original image-prompt-response triplet, enabling vision-oriented preference learning with direct preference supervision (i.e., contrastive response pairs). 
%Specifically, the contrastive response pair comprises two preferred responses ($y_1^w$ and $y_2^w$) corresponding to two contrastive images ($m$ and $m'$). 
Specifically, we design a symmetric preference optimization formulation, which simultaneously maximizes the likelihood of ranking $y_m$ over $y_m'$ given $(m, x)$ and that of ranking $y_m'$ over $y_m$ given $(m', x)$, where $m'$ is  an image semantically similar to $m$ (identified via CLIP visual similarity), and $y_m'$ is the preferred response for $m'$ given $x$. The symmetric optimization compels the model to thoroughly differentiate between contrastive responses under varying visual contexts.   By aligning with standard DPO in terms of using contrastive responses and fixed inputs $(m,x)$, our method naturally eliminates the two partition functions in the reward formulation, yielding a rigorous objective. Further, beyond conventional ordinal preference ranking optimization (i,e., $y_w\succ y_w'$), SymMPO introduces preference margin consistency regularization that enforces consistency in  magnitude of preference gaps between contrastive responses. 
Additionally, SymMPO integrates the established response-oriented preference learning loss due to its remarkable performance in ensuring response quality, and introduces a cost-effective caption-anchored pipeline for constructing response preference pairs, facilitating efficient training and evaluation of our model.

Our main contributions can be summarized in three key aspects:
\begin{itemize}[leftmargin=2em]
    \item We identify two key limitations in existing DPO-based multimodal preference optimization methods for MLLMs: non-rigorous objective function and indirect preference supervision for vision-oriented preference learning. For the non-rigorous objective function, we also provide detailed mathematical derivation in Appendix~\ref{appendix:discussion_of_partition_function}.
   % the inconsistency between practical implementations and theoretical optimization objectives caused by partition function discrepancies, and (2) the indirect approach to vision-oriented preference alignment, which reduces the effectiveness of multimodal alignment of mitigating hallucination.
    \item We propose SymMPO that effectively leverages preferred responses for contrastive images, supporting direct vision-oriented preference learning with a theory-consistent symmetric objective function. Additionally, we design a preference margin consistency regularization to quantitatively constrain the preference gap between symmetric response pairs, enabling more precise contrastive preference learning.  
    \item We empirically validate the superiority of SymMPO over existing approaches through comprehensive experiments on four well-established MLLM hallucination benchmarks, as well as one benchmark designed to evaluate the ability of MLLMs to answer question based on images.
    % The code and data is available at~\href{https://anonymous.4open.science/r/SymMPO-B44F}{https://anonymous.4open.science/r/SymMPO-B44F}.%, demonstrating its superior performance over existing approaches.
\end{itemize}

\section{Preliminary}
%To facilitate the following discussion of the limitations of existing DPO-based multimodal preference optimization methods and derivation of our proposed method, in this section, we first present the Proximal Policy Optimization (PPO) method, which is the theoretical foundation of the DPO method, and then introduce the standard DPO method.   %analysis of DPO-based multimodal preference optimization, 
To facilitate our method presentation, we first review Proximal Policy Optimization (DPO's foundation), standard DPO, and existing multimodal DPO learning paradigm.
% In this section, we establish the theoretical foundation of DPO by presenting its formal derivation from Proximal Policy Optimization (PPO) through the Bradley-Terry model, followed by an explanation of how to adapt it to MLLMs.

% \subsection{Proximal Policy Optimization}
\subsection{RLHF with Bradley-Terry Reward Model}

Reinforcement Learning with Human Feedback (RLHF)~\cite{stiennon2020learning, ouyang2022training} has become a widely adopted approach for aligning LLMs with human preferences and expectations, enabling the generation of responses that better meet user needs. Among the various RLHF methods, Proximal Policy Optimization (PPO)~\cite{schulman2017proximal} stands out as one of the most commonly used methods, primarily due to its stability and efficiency in policy optimization. PPO operates in two main stages: first, it trains a reward model that serves as a proxy for human feedback; second, it optimizes a policy model based on this reward model to guide the LLM toward generating responses that more effectively align with human expectations and preferences.

Specifically, to derive the reward model, PPO employs the widely adopted Bradley-Terry model~\cite{bradley1952rank}, which maximizes the likelihood of the preferred (winning) response $y_w$ being ranked higher than the less preferred (losing) response $y_l$ for a given input $x$. This is formalized as follows:
\begin{equation}
\begin{aligned}
    P_{BT}(y_w\succ y_l\vert x)=\sigma(r_\phi(x,y_w)-r_\phi(x,y_l))=\frac{\exp(r_\phi(x,y_w))}{\exp(r_\phi(x,y_w))+\exp(r_\phi(x,y_l))},
\end{aligned}
\label{Formula:Bradley_Terry_model}
\end{equation}
where $\phi$ represents the parameters of the reward model. $\sigma(\cdot)$ denotes the sigmoid function. %Through this training paradigm, the reward model serves as a proxy for human feedback, enabling the iterative refinement of LLM responses toward better alignment with human expectations and preferences.

Based on the trained reward model $r_\phi$, PPO optimizes the policy model as follows: 
\begin{equation}
    \max_{\pi_\theta}\mathbb{E}_{x\sim \mathcal{D},y\sim\pi_\theta(\cdot\vert x)}\Big[r_\phi(x,y)-\beta D_{KL}\big(\pi_\theta(\cdot\vert x)\Vert\pi_{ref}(\cdot\vert x)\big)\Big],
    \label{Formula:PPO}
\end{equation}
where $x\in\mathcal{D}$ represents the input text sampled from dataset $\mathcal{D}$, and $y$ denotes the response generated by the current policy model $\pi_\theta$ parameterized with $\theta$. $\pi_{ref}$ is a fixed reference model used for 
constraining the policy model through Kullback-Leibler (KL) divergence $D_{KL}(\cdot\Vert\cdot)$.  This objective function ensures the policy model $\pi_\theta$ progressively aligns with human preferences captured by the reward model $r_\phi$, while maintaining generation stability and preventing mode collapse. The hyper-parameter $\beta$ controls the strength of the KL penalty.
%modulates the KL divergence $D_{KL}(\cdot\Vert\cdot)$, which serves as a regularization optimization to constrain the policy model's deviation from the reference model.
% The reward model $r_\phi(x,y)$ evaluates the quality of response $y$ given input $x$, providing a quantitative measure of the performance of the policy model $\pi_\theta$ relative to human preferences.

\subsection{Standard Direct Preference Optimization}
%DPO has emerged as an efficient alternative to traditional Reinforcement Learning from Human Feedback (RLHF) approaches for enhancing the response quality of LLMs. 
%exhibits several limitations~\cite{ivison2024unpacking}, such as: (1) computational complexity due to the separate additional reward model training process, and (2) inefficiency stemming from its multi-step iterative optimization procedure. 
Although the PPO framework demonstrates effective performance in RLHF applications, it involves a separate additional reward model training process, limiting the training efficiency~\cite{ivison2024unpacking}. 
To address this, DPO~\cite{rafailov2023direct} was introduced to eliminate the need for explicit reward modeling, while preserving remarkable effectiveness in aligning LLMs' output with human preferences. In particular, DPO introduces an implicit reward formulation that effectively integrates the probabilities yielded by the policy model and reference model, enabling the direct optimization of the policy model through establishing a direct mapping between policy parameters and human preferences via the Bradley-Terry model.
In particular, the implicit reward formulation  integrates the probabilities yielded by the policy model and reference model as follows:
\begin{equation}
    r(x,y)=\beta\log\frac{\pi_\theta(y\vert x)}{\pi_{ref}(y\vert x)}+\beta\log Z(x),
\label{Formula:implicit_reward}
\end{equation}
where $Z(x)=\sum_y\pi_{ref}(y\vert x)\exp\big(\frac{1}{\beta}r(x,y)\big)$ represents an intractable partition function, and $\beta$ is a hyper-parameter that controls the deviation from the reference model. % as in standard PPO (Equation~\ref{Formula:PPO}).

Although the implicit reward involves an intractable partition function, substituting it into the Bradley-Terry model (Equation~\ref{Formula:Bradley_Terry_model}) and applying the negative logarithmic transformation allows us to derive a DPO objective function free of any intractable terms, as follows:
\begin{equation}
\begin{aligned}
    \mathcal{L}_{DPO}
    % &=-\log\sigma(\beta\log\frac{\pi_\theta(y_w\vert x)}{\pi_{ref}(y_w\vert x)}-\beta\log\frac{\pi_\theta(y_l\vert x)}{\pi_{ref}(y_l\vert x)} \\
    % &\hspace{2.5cm}+\beta\log Z(x)-\beta\log Z(x)) \\
    &=-\mathbb{E}_{(x.y_w,y_l)\sim\mathcal{D}}\Big[\log\sigma\Big(\beta\log\frac{\pi_\theta(y_w\vert x)}{\pi_{ref}(y_w\vert x)}-\beta\log\frac{\pi_\theta(y_l\vert x)}{\pi_{ref}(y_l\vert x)}\Big)\Big],
\end{aligned}
\end{equation}
where, consistent with Equation~\ref{Formula:Bradley_Terry_model}, $y_w$ and $y_l$ denote the preferred and less preferred response, respectively, for given input text $x$.   %The elimination of intermediate reward modeling and multi-step optimization significantly reduces computational complexity while maintaining alignment performance.

%\section{ Discussion of Existing Methods}\label{section:limitation}
%In this section, we first overview the learning paradigm of existing DPO-based multimodal preference optimization methods, which typically involves two key components: response-oriented preference learning and vision-oriented preference learning. % We then identify and analyze its two key limitations.
%applying DPO to MLLMs. Subsequently, we introduce Symmetric Multimodal Preference Optimization (SymMPO), a novel framework designed to address these limitations while enhancing the preference optimization process for MLLMs.
\subsection{Multimodal Direct Preference Optimization}\label{preliminary}
Existing DPO-based multimodal preference optimization methods typically involve two key components: response-oriented preference learning~\cite{sun2023aligning, yu2024rlaif} and vision-oriented preference learning~\cite{wang2024mdpo, yang2025mitigating}. While the former is universally adopted for response preference alignment, the latter is optional for enhancing visual interpretation.

\paragraph{Response-oriented Preference Learning.}
The standard DPO is designed for aligning LLMs to human preferences based on pairs of preferred and less preferred responses (i.e., $y_w$ and $y_l$). To align MLLMs with human preferences, existing methods typically directly extend the DPO objective by including an image condition $m$ as follows, 
%which is shared by both preferred and less preferred samples. 
%This extension preserves the original DPO formulation's mathematical structure while introducing consistent visual content. 
%The modified preference optimization objective for MLLMs can expressed as:
\begin{equation}
\begin{aligned}
    \mathcal{L}_{DPO_m}=-\mathbb{E}_{(x,m,y_w,y_l)\sim\mathcal{D}}\Big[\log\sigma\Big(\beta\log\frac{\pi_\theta(y_w\vert m,x)}{\pi_{ref}(y_w\vert m,x)}-\beta\log\frac{\pi_\theta(y_l\vert m,x)}{\pi_{ref}(y_l\vert m,x)}\Big)\Big].
\label{Formula:DPO_m}
\end{aligned}
\end{equation}
This objective maximizes the likelihood of the preferred (winning) response $y_w$ being ranked higher than the less preferred (losing) response $y_l$ for a given multimodal input ($m$, $x$). %Intuitively, as only the response is the variable between the two terms, 
Intuitively, it expects the model to capture the preference distinctions based on the overall multimodal input. 

\paragraph{Vision-oriented Preference Learning.}
Despite its effectiveness, the overall response-oriented preference learning cannot guarantee MLLMs to accurately interpret the visual content. Therefore, recent studies incorporate a vision-oriented contrastive mechanism to enhance the MLLMs' visual understanding and hence generate more accurate responses with less hallucination. Specifically, in addition to $\mathcal{L}_{DPO_m}$, they introduce a vision-oriented contrastive objective $ \mathcal{L}_{VCO}$ defined as:
\begin{equation}
\begin{aligned}
    \mathcal{L}_{VCO}=-\mathbb{E}_{(x,m_w,m_l,y_w)\sim\mathcal{D}}\Big[\log\sigma\Big(\beta\log\frac{\pi_\theta(y_w\vert m_w,x)}{\pi_{ref}(y_w\vert m_w,x)}-\beta\log\frac{\pi_\theta(y_w\vert m_l,x)}{\pi_{ref}(y_w\vert m_l,x)}\Big)\Big],
\end{aligned}
\label{Formular:vision}
\end{equation}
where $m_w$ represents the original image sampled from training dataset, and $m_l$ denotes a contrastive variant of $m_w$ generated through certain image manipulation operations (e.g., cropping and rotation). As can be seen, unlike  $\mathcal{L}_{DPO_m}$ (see Equation~\ref{Formula:DPO_m}), this objective takes the image-text pairs with fixed text (i.e., $x$) and varied images (i.e., $m_w$ and $m_l$) as the input condition, and adopts the same response (i.e., $y_w$) for both terms.
By making the image condition the only variable, this objective encourages the model to learn preference distinctions based solely on visual information, thereby enhancing the MLLM's understanding of visual input. % Notably, prior work~\cite{wang2024mdpo, xie2024v, fu2025chip} has shown that using contrastive images (i.e., $m_w$ and $m_l$) helps MLLMs grasp subtle differences between images, improving their performance on VQA tasks.

%Consequently, the objective function of current vision-oriented contrastive mechanism is a deviated objective that diverges from the theoretically optimal target. 
%This misalignment between the practical implementation and theoretical derivation prevents existing methods from achieving the theoretically optimal performance. 

% \subsection{Visual-Text Optimization Deviation}

\section{Symmetric Multimodal Preference Optimization}\label{symm}
To address the non-rigorous objective function and indirect preference supervision of existing vision-enhanced multimodal preference optimization methods, we propose SymMPO that conducts symmetric pairwise preference learning and preference margin consistency regularization, to enhance visual understanding and reduce MLLM hallucinations. Notably,  SymMPO also incorporates the response-oriented preference learning, a well-established universal component for ensuring response quality. Here, we omit its details as they are covered in Section~\ref{preliminary}. 
\textbf{Pair-wise Preference Learning.} %To facilitate following presentation, we first define a standard triplet sample $(m, x, y_w)$, where $m$, $x$ and $y_w$ represent the input image, input prompt, and preferred response, respectively.
Unlike existing vision-enhanced multimodal preference alignment  methods that use indirect preference supervision (i.e., contrastive images), we seek direct preference supervision by further generating a preferred response $y_w'$ for the contrastive image $m'$  given the same prompt $x$. $m'$ is designed to differ only subtly from $m$,
thereby leading to similar preferred responses (i.e., $y_w$ and $y_w'$) with minor variations.
This intrinsic relationship naturally establishes $y_w$ and $y_w'$ as a contrastive pair, for enhancing the MLLMs' visual understanding capabilities. Towards comprehensive preference learning, SymMPO designs the following symmetric pairwise reward modeling function with the Bradley-Terry model:
\begin{equation}
\begin{aligned}
    &P_{BT}(y_w\succ y_w'\vert m,x)\wedge P_{BT}(y_w'\succ y_w\vert m',x) \\
    =\ &\sigma\big(r(m,x,y_w)-r(m,x,y_w')\big)\cdot\sigma\big(r(m',x,y_w')-r(m',x,y_w)\big).
\end{aligned}\label{eq9}
\end{equation}
%where %$m$ and $m'$ are paired contrastive images designed to maintain visual similarity while preserving semantically meaningful distinctions. $y_w$ and $y_w'$ are the corresponding responses of the multimodal input $(m, x)$ and $(m', x)$, respectively. 
This function essentially jointly models two pair-wise rewards, each aims to maximize the likelihood that the original preferred response ranks higher than the hard negative (contrastive) response for the given multimodal input. This encourages MLLMs to better interpret the given input image and reduce hallucinations. Then following DPO, we derive the optimization loss by applying the negative logarithmic transformation to the pairwise reward modeling function as follows, 
\begin{equation}
\begin{aligned}
    \mathcal{L}_{Pair} =&-\mathbb{E}_{(x,m,m',y_w,y_w')\sim\mathcal{D}}\Big[\log\sigma\big(r(m,x,y_w)-r(m,x,y_w')\big)+\log\sigma\big(r(m',x,y_w')-r(m',x,y_w)\big)\Big]\\
    =&-\mathbb{E}_{(x,m,m',y_w,y_w')\sim\mathcal{D}}\Big[\log\sigma\Big(\beta\log\frac{\pi_\theta(y_w\vert m,x)}{\pi_{ref}(y_w\vert m,x)}-\beta\log\frac{\pi_\theta(y_w'\vert m,x)}{\pi_{ref}(y_w'\vert m,x)}\Big) \\
    &\hspace{3cm}+\log\sigma\Big(\beta\log\frac{\pi_\theta(y_w'\vert m',x)}{\pi_{ref}(y_w'\vert m',x)}-\beta\log\frac{\pi_\theta(y_w\vert m',x)}{\pi_{ref}(y_w\vert m',x)}\Big)\Big].
\end{aligned}
\label{formula:symmpo}
\end{equation}
Since the above objective function aligns with standard DPO by using response pairs as direct preference supervision, SymMPO naturally cancels out the intractable partition functions that share the same multimodal inputs, leading to a rigorous objective function. %resolving the non-rigorous objective function of existing methods. In other words, this formulation maintains the theoretical consistency of DPO while extending its capabilities to handle multimodal inputs through paired image comparison.

%Unlike the standard multimodal DPO, our SymMPO does not need additional operations, which typically involve LLMs' understanding, for identifying which response should be preferred. 
%Therefore, our method is more efficient, compared to the standard multimodal DPO, leave alone those further involve 

%its requirement for only a single response per image-text pair, which substantially reduces the data generation complexity compared to most previous preference optimization methods that need pairwise response comparisons. This efficiency gain is achieved while maintaining the model's ability to learn effective visual-textual alignments through the innovative use of paired image comparisons and their corresponding responses.

\textbf{Preference Margin Consistency Regularization.} As shown in Equation~\ref{eq9}, the Bradley-Terry model only separately constrains $r(m,x,y_w)>r(m,x,y_w')$ and  $r(m',x,y_w')>r(m',x,y_w)$. However, since $m'$ is a perturbed variant of $m$,  the preference margin between $r(m,x,y_w)$ and $r(m,x,y_w')$ should be similar to that between $r(m',x,y_w')$ and $r(m',x,y_w)$ for near-identical inputs (i.e., $(m,x)$ and $(m',x)$). 
Thus, beyond the traditional ordinal preference learning, we incorporate a preference margin consistency regularization, which quantitatively regulate the response preference margin across contrastive images as follows:
\begin{equation}
 \left\{
\begin{aligned}
    &\mathcal{L}_{Margin}=\mathbb{E}_{(x,m,m',y_w,y_w')\sim\mathcal{D}}\Big(\Delta(m,x,y_w,y_w')-\Delta(m',x,y_w',y_w)\Big)^2, \\
    &\Delta(m,x,y_w,y_w')=r(m,x,y_w)-r(m,x,y_w')=\log\frac{\pi_\theta(y_w\vert m,x)}{\pi_{ref}(y_w\vert m,x)}-\log\frac{\pi_\theta(y_w'\vert m,x)}{\pi_{ref}(y_w'\vert m,x)},
\end{aligned}\right.
\label{formula:regular_term}
\end{equation}
where $\Delta(m,x,y_w,y_w')$ quantifies the preference margin between $y_w$ and $y_w'$ given $(m,x)$.

\begin{comment}
\textcolor{blue}{The visual consistency regularization term is then formulated as:}
\begin{equation}
\begin{aligned}
    \mathcal{L}_{reg}=\Big(\Delta(m,x,y_w,y_w')-\Delta(m',x,y_w',y_w)\Big)^2.
\end{aligned}
\end{equation}
\end{comment}
%\textcolor{blue}{This regularization term enforces consistent reward margins across symmetric image-response pairs, promoting more stable and reliable preference learning. The squared difference penalizes discrepancies in the model's relative preference judgments when the visual context is reversed, thereby strengthening the alignment between visual understanding and response generation.}

Moreover, to prevent the model from reducing the likelihood of the preferred response for optimizing the likelihood gap between preferred and less preferred responses, which can be harmful to preference alignment, we also adopt the anchored preference regularization~\cite{wang2024mdpo, yang2025mitigating} as follows:
\begin{equation}
\begin{aligned}
    \mathcal{L}_{AncPO}=-\mathbb{E}_{(x,m,m',y_w,y_w')\sim\mathcal{D}}\Big[\log\sigma\Big(\beta\log\frac{\pi_\theta(y_w\vert m,x)}{\pi_{ref}(y_w\vert m,x)}-\delta\Big)+\log\sigma\Big(\beta\log\frac{\pi_\theta(y_w'\vert m',x)}{\pi_{ref}(y_w'\vert m',x)}-\delta\Big)\Big],
    \label{Formula:anchor}
\end{aligned}
\end{equation}
%where $\beta$ serves as the hyper-parameter controls the deviation from the reference model (consistent with conventional DPO formulations), and 
where $\delta$ is the anchor that encourages the model to yield high likelihoods for preferred responses. %, ensuring they maintain minimum likelihood thresholds during optimization.

Ultimately, the complete optimization objective of SymMPO is formally defined as follows:
\begin{equation}
\begin{aligned}
    \mathcal{L}_{SymMPO}=\mathcal{L}_{DPO_m}+\lambda\mathcal{L}_{Pair}+\gamma\mathcal{L}_{Margin}+\eta\mathcal{L}_{AncPO},
    \label{Formula:final_loss}
\end{aligned}
\end{equation}
where $\lambda$, $\gamma$, and $\eta$ are weighting hyper-parameters. 

%Essentially, different from mainstream multimodal DPO with visual-oriented contrastive objectives (see Equation~\ref{Formula:DPO_m}), in addition to the response for the original image, SymMPO also requires a response based on the contrastive image. Nevertheless, a particularly noteworthy advantage of SymMPO is its ability to directly leverage the natural relationship between the image and the response to identify the preferred and less preferred responses. Moreover, our SymMPO effectively conducts vision-oriented preference learning within the pairwise preference learning objective using contrastive image pairs, which standard multimodal DPO fails to achieve.

\begin{figure}
    \centering
    \includegraphics[width=1.0\linewidth]{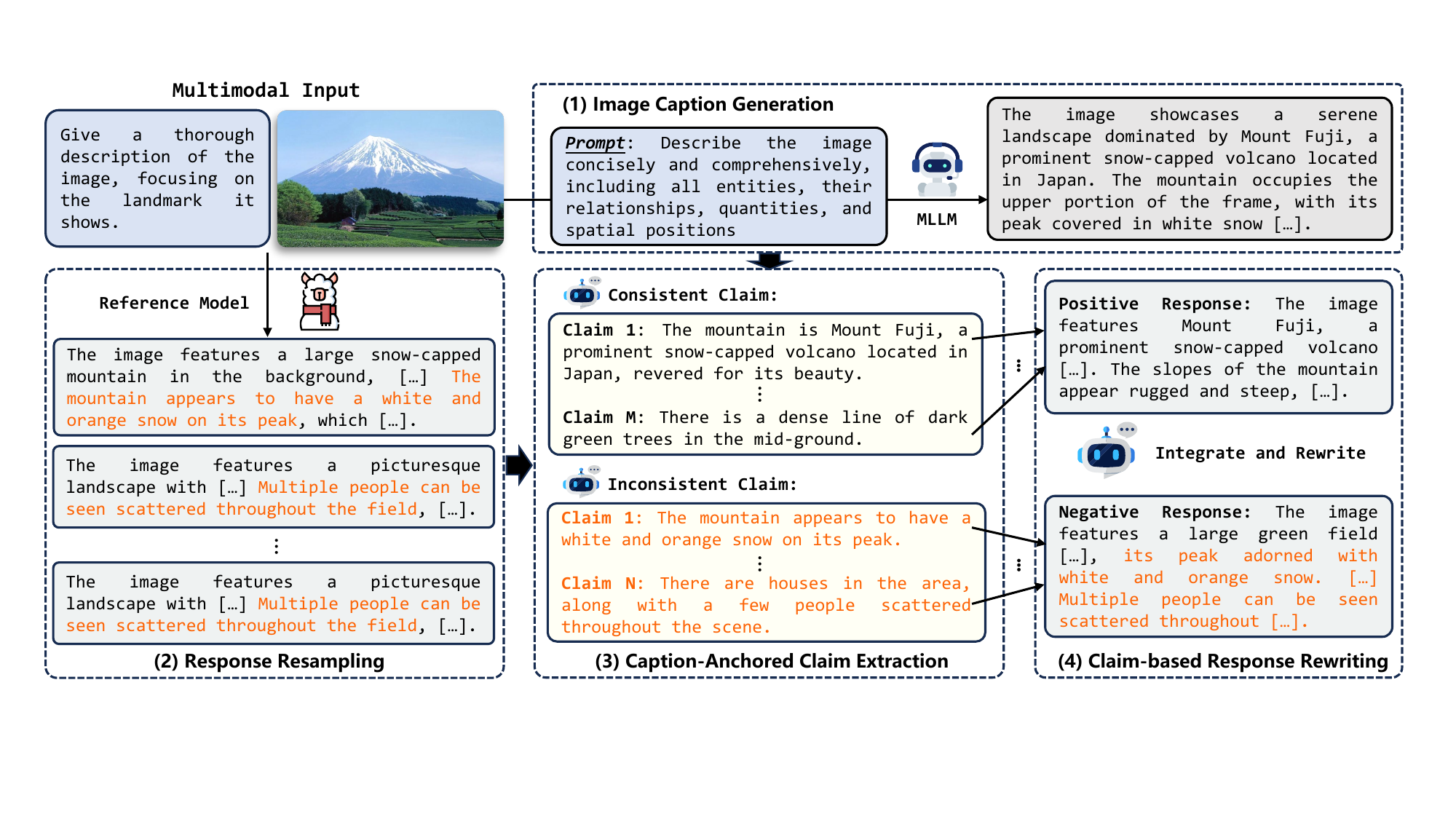}
    \caption{The caption-anchored response preference pair construction pipeline. %: (1) Image caption generation using an open-source MLLM. (2) Response resampling from the reference model. (3) Caption-anchored claim extraction using an LLM. (4) Claim-based response rewriting using an LLM.
    }
    \label{fig:data_construction}
    \vspace{-1em}
\end{figure}

\section{Experiment}\label{section:experiment}
In this section, we present our training data construction process, experimental setup and results.% analysis.%then  provide a comprehensive evaluation to assess the effectiveness of SymMPO in mitigating hallucination.

\subsection{Training Data Construction}\label{data_construction}
%\paragraph{Training Data.}
In this work, we adopt the same set of 21.4k image-prompt pairs from TPO~\cite{he2024topic}, which aggregates multiple public datasets, including VQA v2~\cite{goyal2017making}, MSCOCO~\cite{lin2014microsoft} and TextVQA~\cite{singh2019towards}.
To construct preference data for model optimization, we first generate contrastive responses $(y_w, y_l)$ for each image-prompt pair $(m,x)$ to compute $\mathcal{L}_{DPO_m}$. Subsequently, we create a contrastive image $m'$ with its corresponding positive response $y_m'$, forming contrastive triplets $(m, x, y_m)$ and $(m', x, y_m')$ for calculating $\mathcal{L}_{Pair}$, $\mathcal{L}_{Margin}$, and $\mathcal{L}_{AncPO}$.  
This process essentially involves two key modules: preference response pair construction and contrastive image construction, while $y_m'$ for $m'$ can be easily obtained by the preference response pair construction module.

%\subsection{Caption-Anchored Response Preference Pair Construction Pipeline}\label{data_construction}
%%\paragraph{Training Data} 
%As shown in Equation~\ref{Formula:final_loss}, there are four loss terms are used for model optimization. In particular, the commonly used loss $\mathcal{L}_{DPO_m}$ requires a positive response and a negative response for each multimodal input; while $\mathcal{L}_{Pair}$, $\mathcal{L}_{Margin}$, and $\mathcal{L}_{AncPO}$ require a positive response for the original  multimodal input and a positive response for a slightly varied multimodal input with a contrastive image. Essentially, we need to 

%typically fall into two categories. (1) Direct Correction via Commercial MLLMs. This approach employs commercial, utilizing commercial state-of-the-art  MLLMs such as GPT-4V~\cite{openai2023gpt4v} to rectify errors in reference model's outputs, where the original outputs serve as negative responses and the rectified versions as positive responses. While effective, this approach incurs substantial API costs associated with commercial MLLM usage.

%Therefore, to support the model training,  we first resort to the claim extraction-based response generation paradigm~\cite{yu2024rlaif, he2024topic} to construct positive-negative response preference pairs rather than the direct response correction-based approach~\cite{zhou2024aligning, sarkar2024data, xiao2024detecting}, which relies on costly API calls to commercial MLLMs (e.g. GPT-4V~\cite{openai2023gpt4v}).
\textbf{Preference Response Pair Construction}.
%Similar to existing DPO-based methods~\cite{wang2024mdpo, xie2024v, fu2025chip}, we involve the loss $\mathcal{L}_{DPO_m}$ for model optimization, which requires both positive and negative responses for each multimodal input.  
While previous works have proposed various methods for preference pair construction~\cite{zhao2023beyond, yu2024rlaif}, these approaches typically suffer from either costly API calls to commercial MLLMs (e.g., GPT-4V~\cite{openai2023gpt4v}) or substantial computational overhead with complex multi-stage processing. %, such as claim extraction, LLM-based claim-to-question conversion and MLLM-based trustworthy verification for each claim. 
Therefore, we design a cost-effective Caption-Anchored Claim Extraction-and-Rewriting pipeline, as illustrated in Figure~\ref{fig:data_construction}, comprising four efficient stages: 1) Image Caption Generation, 2) Response Resampling, 3) Caption-Anchored Claim Extraction, and 4) Claim-Based Response Rewriting. First, we use an open-source MLLM (e.g., Qwen2.5-VL-32B~\cite{bai2025qwen2, wang2024qwen2}) to generate a detailed caption for the image. Next, multiple responses are generated from the reference model. Unlike prior approaches~\cite{yu2024rlaif, he2024topic}, the third stage of our pipeline directly extracts atomic claims by performing fine-grained comparison between each response and the detailed image caption, leveraging a high-performance LLM (e.g., DeepSeek-V3~\cite{liu2024deepseek}). This simplifies the process by removing the need for explicit claim extraction, claim-to-question conversion, and per-claim verification as required in previous methods~\cite{yu2024rlaif, he2024topic}. Instead, our approach only requires generating the image caption and designing a prompt that instructs the LLM to extract positive claims (consistent with the caption) and negative claims (inconsistent with it). Finally, an LLM (e.g., DeepSeek-V3) is further used as a rewrite model to generate positive and negative responses based on the extracted consistent and inconsistent claims, respectively.

\textbf{Contrastive Image Construction}. Unlike existing methods that generate the contrastive image $m'$ by applying transformations or noise to the original image $m$, we adopt an alternative construction approach.  Specifically, we compute CLIP~\cite{radford2021learning}  embeddings for all images in the dataset and pair each image with its nearest neighbor based on cosine similarity. This ensures $m'$ and $m$ share high visual similarity while exhibiting subtle differences, making their positive responses ($y_m, y_m'$) more challenging to distinguish, thereby enhancing symmetric pair-wise preference learning.  %Crucially, this approach preserves the authenticity of $m'$ as it is selected from real-world samples rather than artificially constructed. 

% We believe that using CLIP can make $m'$ share high visual similarity but subtle differences with $m$, leading to their corresponding positive responses ($y_m, y_m'$) also highly similar and hard to be differentiated.  Specifically, we first leverage CLIP to compute image embeddings for all images in the dataset, and then pair each image with its most similar counterpart in the dataset based on their cosine similarities. %These derived contrastive images are combined with the same prompts of original images, forming new image-prompt pairs. We then employ the aforementioned caption-anchored response preference pair construction pipeline to generate the final training dataset. Specifically, for each original image-prompt pair, we generate both positive and negative responses, enabling the calculation of $\mathcal{L}_{DPO_m}$. For each newly derived image-prompt pair (with a contrastive image), we generate only a positive response participating the calculation of $\mathcal{L}_{Pair}$.

\subsection{Experimental Setup.}\label{section:experimental_setup}
%To comprehensively evaluate SymMPO's effectiveness in mitigating hallucination in MLLMs, 

\textbf{Benchmarks.} For evaluation, we adopt five established benchmarks: 1) \textbf{HallusionBench}~\cite{guan2024hallusionbench} evaluates both language hallucination and visual illusion, featuring 346 images and 1,129 prompts. It employs GPT-4~\cite{achiam2023gpt} to compare MLLM outputs against ground truth, using three evaluation metrics: question-level accuracy (qAcc), figure-level accuracy (fAcc), and overall accuracy across all prompts (aAcc). For cost efficiency, we substitute the original GPT-4~\cite{achiam2023gpt} evaluator with DeepSeek-V3~\cite{liu2024deepseek}.
   2)  \textbf{Object-HalBench}~\cite{rohrbach2018object} is a benchmark for evaluating common object hallucination in detailed image descriptions. Following~\cite{yu2024rlhf} and~\cite{yu2024rlaif}, we use eight diverse prompts across 300 instances to ensure robust evaluation, reporting both response-level and mention-level hallucination rates.
    3) \textbf{MMHal-Bench}~\cite{sun2023aligning} evaluates response informativeness and hallucination rates using GPT-4 to compare model outputs against human annotations for 96 images. The information score and hallucination rate serve as the evaluation metrics.
    4) \textbf{AMBER}~\cite{wang2023llm} provides 15k fine-grained annotations with well-designed prompts, enabling comprehensive evaluation across three key aspects: object existence, attribute accuracy, and relational correctness. We report the accuracy and F1 score for this discriminative evaluation.
    5) \textbf{MMStar}~\cite{chen2024we} provides 1.5k image-prompt pairs for evaluating six core capabilities and 18 specific aspects of MLLMs, with results summarized in an overall performance score.
 %Through this process, we construct the final SymMPO training dataset.
% Then on this comprehensive dataset, we apply our data construction pipeline to generate high-quality preference response pairs for MLLM training and evaluation.
% \textcolor{red}{ based on method description, you need y1 and y2. So the prompt is how to apply your data construction pipeline to derive y1 and y2? revise here}

\textbf{Baselines.} For fair evaluation, we first adopt the following PPO/DPO-based baselines with publicly available pretrained weights: PPO-based method (LLaVA-RLHF~\cite{sun2023aligning}) and several DPO-based methods, including RLHF-V~\cite{yu2024rlhf} (trained on Muffin-13B~\cite{yu2023reformulating}), as well as 
% DPO~\cite{rafailov2023direct}, mDPO~\cite{wang2024mdpo}, 
POVID~\cite{zhou2024aligning}, HALVA~\cite{sarkar2024data}, HA-DPO~\cite{zhao2023beyond}, RLAIF-V~\cite{yu2024rlaif}, TPO~\cite{he2024topic}, OPA-DPO~\cite{yang2025mitigating}, and HSA-DPO~\cite{xiao2024detecting} (all trained on LLaVA-1.5~\cite{liu2023visual}). 
To eliminate confounding factors from differing experimental settings and ensure  a rigorous comparison, we also incorporate standard DPO~\cite{rafailov2023direct} and mDPO~\cite{wang2024mdpo} in our evaluation. Both methods are trained under the same experimental conditions as SymMPO, including identical training data, parameter configurations, and environment. All models (DPO, mDPO, and SymMPO) for rigorous comparison are trained for 2 epochs with a learning rate of 5e-6 and batch size of 64, using the following hyper-parameters: $\beta=0.1$, $\delta=0$, $\lambda=0.5$, $\gamma=1e-4$, and $\eta=1.0$. The training is performed on 4 NVIDIA A100-40GB~GPUs. 
\begin{table}
  \caption{Main experimental results. The best and second-best results under the same experiment setting are highlighted in boldface and underlined, respectively. %We evaluated SymMPO using both LLaVA-1.5-7B and 13B variants. 
 }
  \label{table:main_results}
  \centering
  \begin{adjustbox}{width=\textwidth}
      \begin{tabular}{l*{12}{c}}
        \toprule
        \multirow{2}{*}{Model} & \multirow{2}{*}{Data Size} & \multirow{2}{*}{Feedback} & \multicolumn{3}{c}{HallusionBench} & \multicolumn{2}{c}{Object-HalBench} & \multicolumn{2}{c}{MMHal-Bench} & \multicolumn{2}{c}{AMBER} & MMStar \\ 
        \cmidrule(r){4-6} \cmidrule(r){7-8} \cmidrule(r){9-10} \cmidrule(r){11-12} \cmidrule(r){13-13}
        & & & qAcc$\uparrow$ &fAcc$\uparrow$ &aAcc$\uparrow$ & Resp.$\downarrow$ & Ment.$\downarrow$ & Score$\uparrow$ & Hall$\downarrow$ & Acc$\uparrow$ & F1$\uparrow$ & Overall$\uparrow$ \\
        \midrule
        \textbf{Muffin-13B}~\cite{yu2023reformulating}  & \ding{55}  & \ding{55} & 6.15 & 12.71 & 41.89 & 53.0 & 24.3 & 2.06  & 66.7 & 74.2 & 80.0 & 25.4  \\
        +RLHF-V~\cite{yu2024rlhf} & 1.4k  & Human & 9.67 & 13.87 & 45.79 & 8.5 & 4.9 & 2.60 & 56.2 & 82.0 & 86.7 & 31.0 \\
        \midrule
        \textbf{LLaVA-1.5-7B}~\cite{liu2023visual} & \ding{55} & \ding{55} & 3.95 & 11.56 & 41.71 & 56.5 & 27.9 & 2.26 & 56.2 & 71.8 & 74.5 & 33.3 \\
        +LLaVA-RLHF~\cite{sun2023aligning} & 122k & Self-Reward & 5.49 & 12.13 & 38.26 & 55.4 & 27.3 & 2.00 & 66.7 & 68.7 & 74.7 & 31.4 \\
        +POVID~\cite{zhou2024aligning} & 17k & GPT-4V & 7.03 & 9.53 & 43.31 & 35.9 & 17.3 & 2.28 & 56.2 & 78.6 & 81.9 & 34.4 \\
        +HALVA~\cite{sarkar2024data} & 21.5k & GPT-4V & 5.49 & 11.27 & 42.42 & 49.1 & 24.6 & 2.14 & 60.4 & 78.0 & 83.5 & 32.3 \\
        +HA-DPO~\cite{zhao2023beyond} & 6k & GPT-4 & 5.49 & 11.56 & 42.16 & 44.9 & 21.8 & 1.97 & 61.5 & 74.2 & 78.0 & 32.6 \\
        +RLAIF-V~\cite{yu2024rlaif} & 74.8k & LLaVA-Next & 5.93 & 5.49 & 36.75 & 9.9 & 4.9 & 3.04 & 39.6 & 72.7 & 84.4 & 34.6 \\
        +TPO~\cite{he2024topic} & 21.4k & LLaVA-Next & 7.03 & 11.27 & 41.62 & 5.0 & 4.7 & 2.76 & 42.7 & 82.2 & 87.2 & 34.2 \\
        +OPA-DPO~\cite{yang2025mitigating} & 4.8k & LLaVA-Next & 6.37 & 11.84 & 42.69 & 6.1 & 3.7 & 2.83 & 46.9 & 81.3 & 85.6 & 33.1 \\
        \rowcolor{blue!15}
        \textbf{+DPO}~\cite{rafailov2023direct} & 21.4k & DeepSeek-V3 & \textbf{7.25} & 7.80 & 40.21 & \textbf{12.9} & \textbf{8.8} & 2.44 & \underline{49.0} & 71.3 & 82.6 & 33.4 \\
        \rowcolor{blue!15}
        \textbf{+mDPO}~\cite{wang2024mdpo} & 21.4k & DeepSeek-V3 & \underline{6.81} & \underline{9.53} & \underline{42.78} & 19.9 & 10.1 & \underline{2.71} & 50.0 & \underline{80.6} & \underline{86.3} & \underline{34.2} \\
        \rowcolor{blue!15}
        \textbf{+SymMPO (Ours)} & 21.4k & DeepSeek-V3 & \textbf{7.25} & \textbf{13.58} & \textbf{44.28} & \underline{19.5} & \underline{9.7} & \textbf{2.89} & \textbf{42.7} & \textbf{82.6} & \textbf{87.7} & \textbf{34.8} \\
        \midrule
        \textbf{LLaVA-1.5-13B}~\cite{liu2023visual} & \ding{55} & \ding{55} & 6.59 & 9.53 & 43.48 & 51.2 & 25.1 & 2.16 & 59.4 & 71.3 & 73.2 & 33.1 \\
        +LLaVA-RLHF~\cite{sun2023aligning} & 122k & Self-Reward & 8.57 & 10.11 & 43.48 & 45.3 & 21.5 & 2.15 & 66.7 & 79.7 & 83.9 & 33.5 \\
        +HALVA~\cite{sarkar2024data} & 21.5k & GPT-4V & 8.79 & 10.11 & 42.24 & 47.0 & 22.9 & 2.30 & 57.3 & 82.9 & 86.5 & 33.1 \\
        +HSA-DPO~\cite{xiao2024detecting} & 8k & GPT-4/4V & 6.15 & 8.95 & 41.62 & 5.4 & 2.9 & 2.55 & 50.0 & 79.8 & 82.8 & 33.7 \\
        +OPA-DPO~\cite{yang2025mitigating} & 4.8k & LLaVA-Next & 6.81 & 12.13 & 42.60 & 7.7 & 4.4 & 3.05 & 38.5 & 84.1 & 87.5 & 32.3 \\
        \rowcolor{blue!15}
        \textbf{+DPO}~\cite{rafailov2023direct} & 21.4k & DeepSeek-V3 & \underline{10.32} & \underline{10.69} & 39.50 & \textbf{15.4} & \textbf{8.5} & 2.65 & 45.8 & 69.2 & 84.6 & 33.0 \\
        \rowcolor{blue!15}
        \textbf{+mDPO}~\cite{wang2024mdpo} & 21.4k & DeepSeek-V3 & 9.23 & \underline{10.69} & \underline{39.85} & 20.9 & 10.8 & \underline{2.93} & \underline{43.8} & \underline{83.8} & \underline{88.8} & \underline{35.0} \\
        \rowcolor{blue!15}
        \textbf{+SymMPO (Ours)} & 21.4k & DeepSeek-V3 & \textbf{10.54} & \textbf{10.98} & \textbf{44.55} & \underline{20.4} & \underline{10.0} & \textbf{3.01} & \textbf{39.6} & \textbf{84.9} & \textbf{89.1} & \textbf{35.2} \\
        % \midrule
        % \rowcolor{gray!15}
        % GPT-4V~\cite{openai2023gpt4v} & - & - & 28.79 & 39.88 & 65.28 & 13.6 & 7.3 & 3.49 & 28.1 & 83.4 & 87.4 & 50.4 \\
        \bottomrule
      \end{tabular}
  \end{adjustbox}
\end{table}

\subsection{Main Results}\label{mainresults}
%Notably, among baseline methods, only DPO and mDPO share the same experimental setting (e.g., training data, parameter configuration, and experimental environment) as ours, enabling valid comparison, while the others do not.
%parameter configuration, and experimental environment, 
%variations in involve Our comparative analysis therefore focuses on DPO, mDPO, and SymMPO, which were trained under identical conditions.
Table~\ref{table:main_results} presents the performance comparison across five benchmarks with two model variants of LLaVA-1.5~\cite{liu2023visual}: LLaVA-1.5-7B and LLaVA-1.5-13B
. 
From this table, we have following observations: (1) SymMPO outperforms existing methods across most evaluation metrics for both LLaVA-1.5-7B and LLaVA-1.5-13B, demonstrating its effectiveness in reducing hallucination in MLLMs. %  Beyond the global results, our analysis reveals several noteworthy findings across individual benchmarks: 
(2) The blue-highlighted rigorous comparisons show SymMPO's consistent superiority over DPO and mDPO across four benchmarks: HallusionBench, MMHal-Bench, AMBER, and MMStar. This advantage stems from our novel vision-oriented contrastive learning framework, which effectively leverages symmetric comparisons through a rigorous objective function derived from standard DPO, thereby enhancing the model's visual understanding.  %  In contrast to mDPO, SymMPO maintains theoretical consistency by %by processing identical input images during optimization, enhancing cross-modal alignment and yielding superior visual contrastive learning outcomes. 
(3) Surprisingly, both mDPO and our SymMPO underperform DPO on Object-HalBench.
This may stem from a misalignment between our preference data construction method and Object-HalBench's evaluation task. In our data construction pipeline, we use an open-source MLLM to generate image captions, which serve as references for extracting consistent/inconsistent claims and generating positive/negative responses. However, these captions tend to focus more on object detection and scene overview rather than fine-grained visual details, thereby introducing noise about subtle visual properties in preference response pairs.
While this noise has a limited impact on benchmarks that assess hallucination using straightforward discriminative questions (e.g., multiple-choice or yes/no questions), it poses a greater challenge for Object-HalBench, which directly evaluates response- and mention-level hallucination in detailed MLLM-generated descriptions. Despite this, SymMPO still outperforms mDPO, demonstrating stronger performance in mitigating hallucination in fine-grained descriptions.

\begin{table}
  \caption{Ablation studies with LLaVA-1.5-7B.}
  \label{table:ablation_results}
  \centering
  \begin{adjustbox}{width=\textwidth}
      \begin{tabular}{l*{10}{c}}
        \toprule
        \multirow{2}{*}{Model} & \multicolumn{3}{c}{HallusionBench} & \multicolumn{2}{c}{Object-HalBench} & \multicolumn{2}{c}{MMHal-Bench} & \multicolumn{2}{c}{AMBER} & MMStar \\ \cmidrule(r){2-4} \cmidrule(r){5-6} \cmidrule(r){7-8} \cmidrule(r){9-10} \cmidrule(r){11-11}
        & qAcc$\uparrow$ &fAcc$\uparrow$ &aAcc$\uparrow$ & Resp.$\downarrow$ & Ment.$\downarrow$ & Score$\uparrow$ & Hall$\downarrow$ & Acc$\uparrow$ & F1$\uparrow$ & Overall$\uparrow$ \\
        \midrule
        \textbf{SymMPO} & \textbf{7.25} & \textbf{13.58} & \underline{44.28} & \underline{19.5} & \textbf{9.7} & \textbf{2.89} & \textbf{42.7} & \textbf{82.6} & \textbf{87.7} & \underline{34.8} \\
        \midrule
        w/o-$\mathcal{L}_{Pair}$ & 6.59 & \underline{11.84} & 43.22 & \textbf{18.1} & \underline{10.6} & \underline{2.53} & \underline{50.0} & 81.7 & 87.1 & 33.8 \\
        w/o-$\mathcal{L}_{Margin}$ & \underline{7.03} & 10.98 & \textbf{44.46} & 21.1 & 11.0 & 2.40 & 54.2 & \underline{82.0} & 87.3 & 34.5 \\
        w/o-$\mathcal{L}_{AncPO}$ & 6.81 & \underline{11.84} & 40.83 & 21.6 & 11.6 & 2.39 & 59.4 & 79.5 & \underline{87.4} & \textbf{36.2} \\
        \bottomrule
      \end{tabular}
  \end{adjustbox}
\end{table}

\subsection{Ablation Study}  

%removing the paired-image contrastive loss (Equation~\ref{formula:symmpo}); (2) \textbf{w/o Reg Term}: removing the regularization loss (Equation~\ref{formula:regular_term}); and (3) \textbf{w/o Anchor}: removing the anchor loss (Equation~\ref{Formula:anchor}). 
To evaluate the contribution of each key component in SymMPO, we introduce three variants: w/o-$\mathcal{L}_{Pair}$, w/o-$\mathcal{L}_{Margin}$, and w/o-$\mathcal{L}_{AncPO}$, where the pair-wise preference loss, the margin consistency regularization, and the anchored preference regularization are removed, respectively. Table~\ref{table:ablation_results} presents the ablation results across five benchmarks using LLaVA-1.5-7B as the backbone. As we can see, the complete SymMPO model outperforms all three variants on most metrics, demonstrating each component's essential contribution to SymMPO's effectiveness. Specifically, the results confirm: (1) the critical importance of pair-wise preference learning for eliminating MLLM hallucination; (2) the necessity of regulating quantitative preference margins between paired samples beyond qualitative preference alignment; and (3) the positive effect of anchored preference regularization in improving training stability by preventing log probability decline of positive responses during optimization.

\begin{figure}
    \centering
    \subfigure[Original]{
        \includegraphics[width=0.13\linewidth]{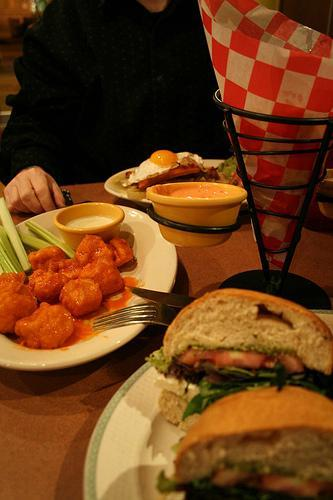}
    }
    \hfill
    \subfigure[Similar]{
        \includegraphics[width=0.2\linewidth]{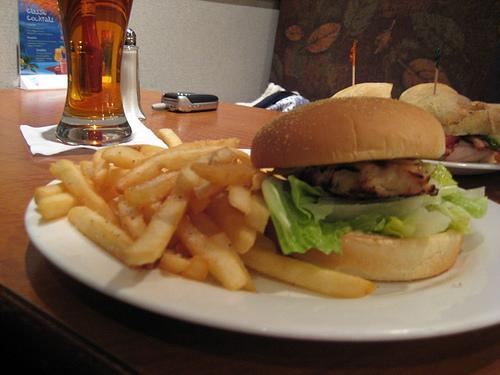}
    }
    \hfill
    \subfigure[Black]{
        \includegraphics[width=0.13\linewidth]{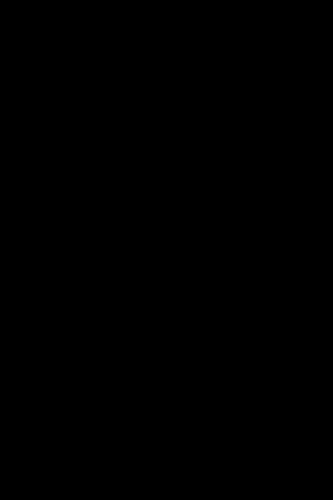}
    }
    \hfill
    \subfigure[Cropped]{
        \includegraphics[width=0.13\linewidth]{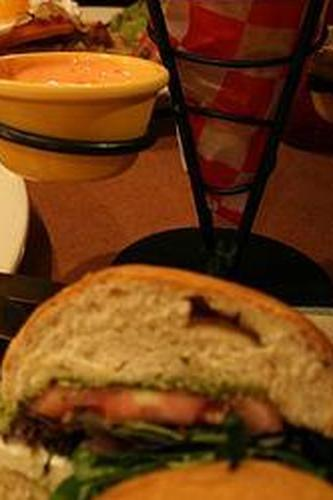}
    }
    \hfill
    \subfigure[Noisy]{
        \includegraphics[width=0.13\linewidth]{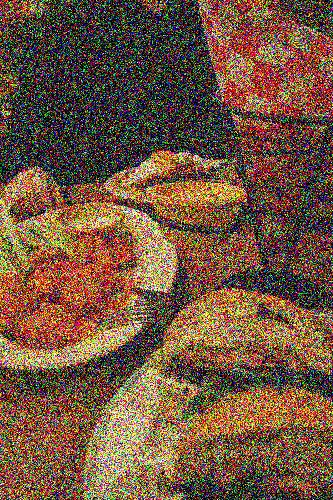}
    }
    \hfill
    \subfigure[Synthetic]{
        \includegraphics[width=0.13\linewidth]{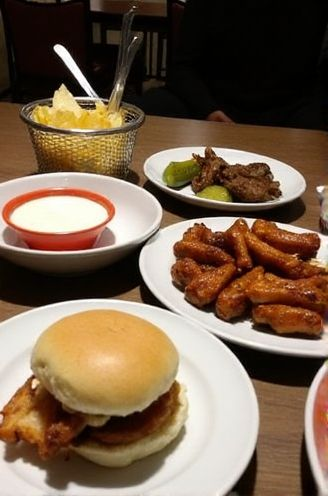}
    }
  %  \vspace{-0.5em}
    \caption{Samples of the original image and its related contrastive images.}
    \label{fig:image_sample}   \vspace{-1em}

\end{figure}

\begin{figure}
    \centering
    \subfigure[Object-HalBench]{
        \includegraphics[width=0.315\linewidth]{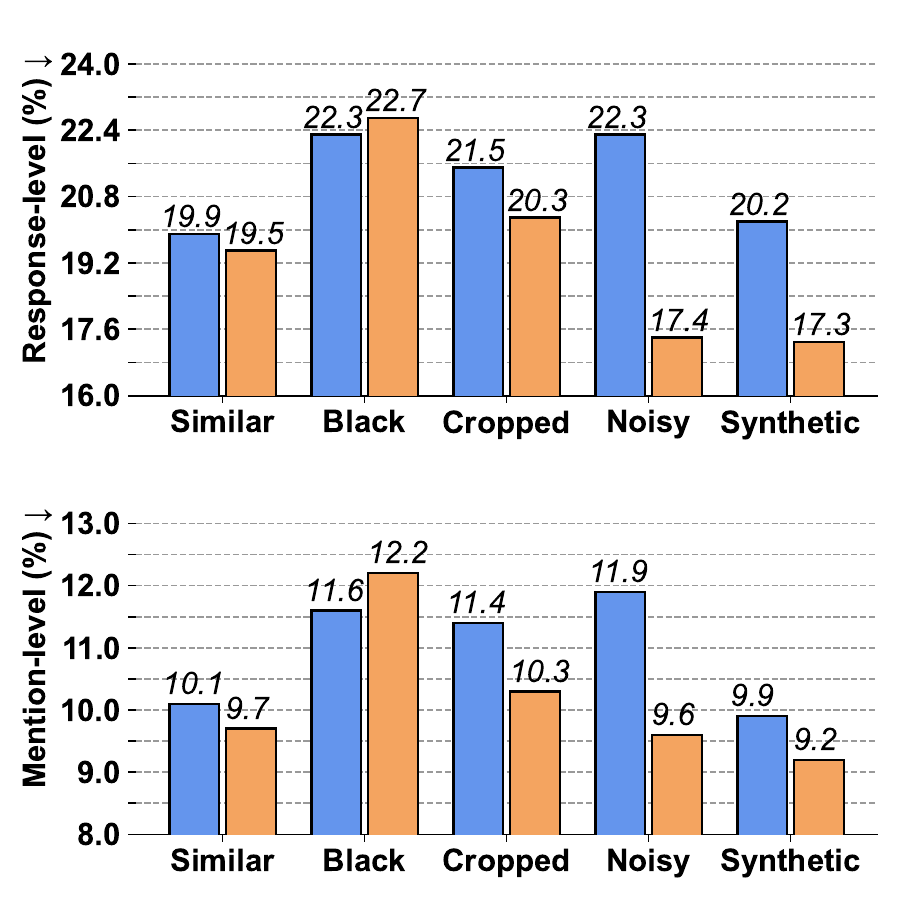}
    }
    % \hfill
    \subfigure[MMHal-Bench]{
        \includegraphics[width=0.315\linewidth]{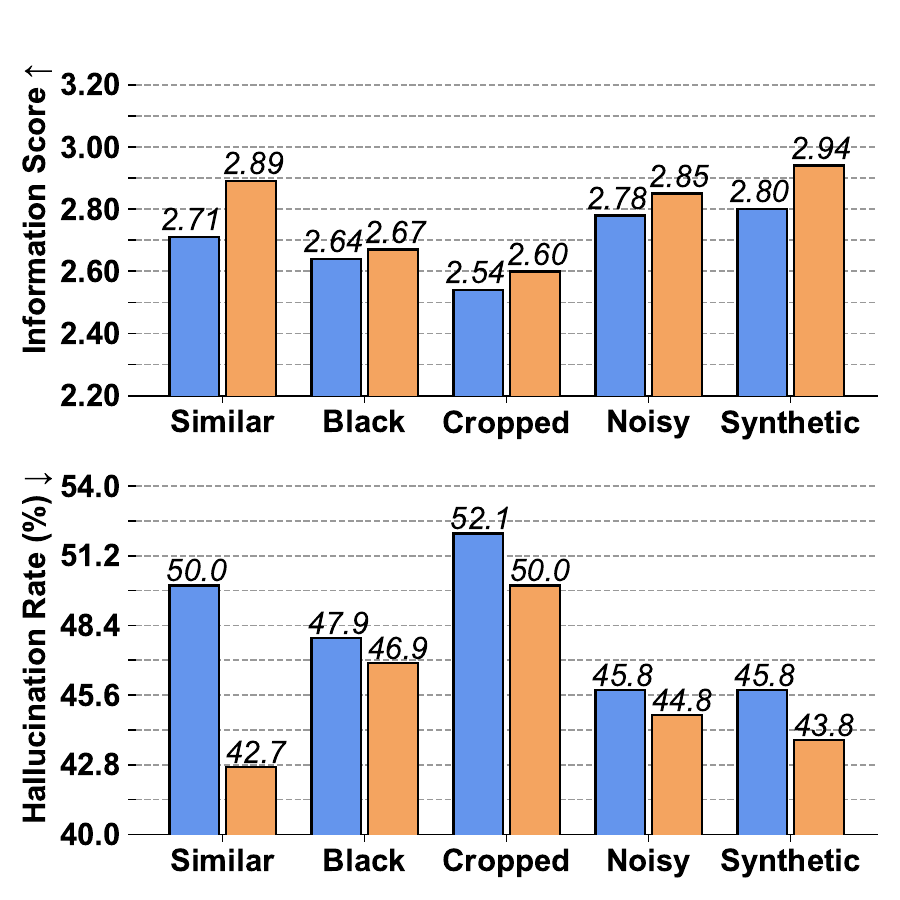}
    }
    % \hfill
    \subfigure[AMBER]{
        \includegraphics[width=0.315\linewidth]{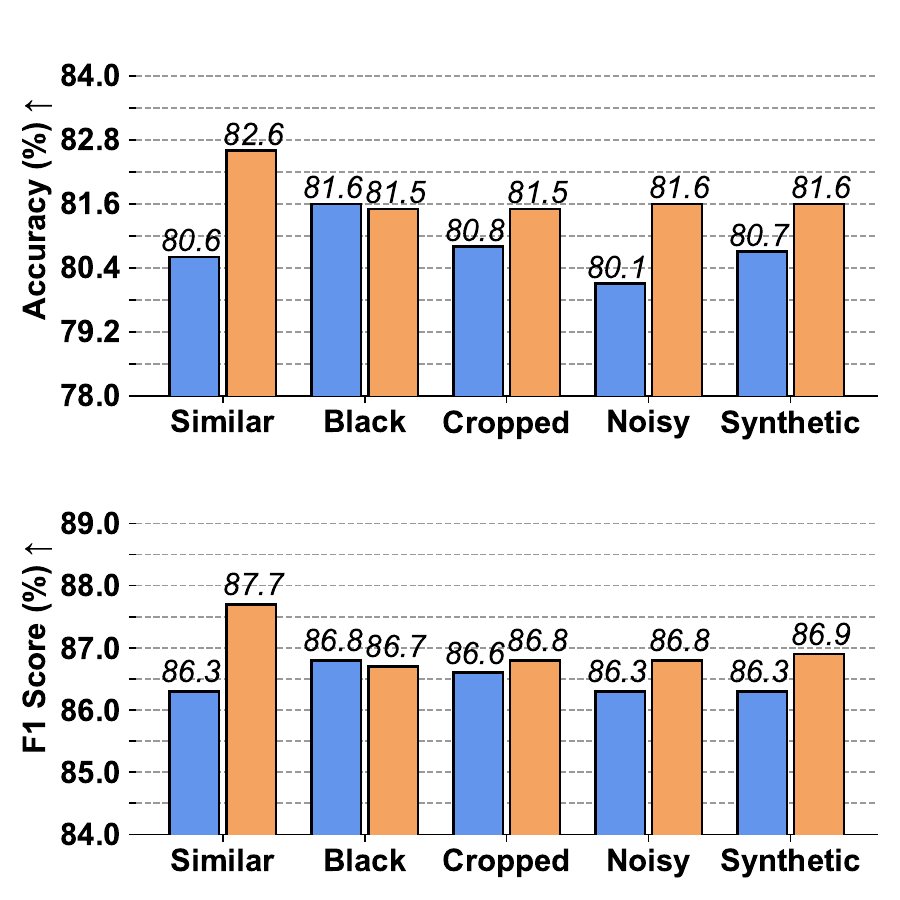}
    }
    % \hfill
  %  \vspace{-0.5em}
    \caption{Results of SymMPO and mDPO using different types of contrastive images ($\uparrow$/$\downarrow$: higher/lower is better). %On Object-HalBench, lower hallucination rates at both response and mention levels indicate better performance; on MMHal-Bench, higher information scores coupled with lower hallucination rates reflect superior results; and on AMBER, higher Accuracy and F1 Scores signify stronger performance. 
    \textbf{\textcolor{orange}{Orange}} represents SymMPO, and \textbf{\textcolor{blue}{blue}} represents mDPO.}
    \label{fig:results_of_image_variant} \vspace{-1em}
\end{figure}

\subsection{Impact of Contrastive Image}
%In the main experiment, we use a semantic similar image as the contrastive image for our symmetric multimodal preference optimization. 
To investigate the impact of contrastive images on our SymMPO, we also perform experiments on four alternative types of contrastive images using LLaVA-1.5-7B. (1) \textbf{Black}: Completely black images containing no visual information; 
(2) \textbf{~Cropped}: Sub-images randomly cropped from original images, with half the original width and height; (3) \textbf{Noisy}: Original images corrupted with Gaussian noise ($\sigma=0.8$); 
%Images with 50\% of both width and height randomly cropped from the original images; 
and (4) \textbf{Synthetic}: Inspired by the great success of image captioning and generation models, we synthesize the contrastive images by FLUX.1-dev~\cite{flux2024} based on the LLaVA-1.5-7B-generated captions of original images. Figure~\ref{fig:image_sample} shows contrastive image examples, where ``Similar'' refers to our CLIP-based contrastive images used in main experiments.   
%textual description  with the standardized prompt: ``Generate a brief and accurate description of the content in the image.'', and then synthesizing images based on these descriptions through FLUX.1-dev~\cite{flux2024}. The representative samples of these image variants are displayed in Figure~\ref{fig:image_sample}. 
%optimizing its two interconnected Bradley-Terry models. 

%To gain deep insights, we compare our SymMPO with mDPO on three mainstream benchmarks, including Object-HalBench, MMHal-Bench, and AMBER. The results of the two models are presented in 
Figure~\ref{fig:results_of_image_variant} shows performance comparison between our SymMPO and mDPO with different types of contrastive images on three mainstream benchmarks, including Object-HalBench, MMHal-Bench, and AMBER. From this figure, we make the following key observation: (1) SymMPO consistently outperforms mDPO across all types of contrastive images, except for black images. This demonstrates the robustness of our model in handling diverse contrastive inputs. The performance degradation of our SymMPO with black images may stem from their lack of visual information, which results in non-meaningful target responses. With insufficient meaningful signals, SymMPO struggles to perform effective preference optimization.
In contrast, mDPO only relies on the varying visual inputs with the same response of  the original image for visual understanding enhancement, making it less susceptible to this limitation. 
(2) Using Similar, Noisy or Synthetic images, demonstrates superior performance over using Black and Cropped images. This possible explanation is that the former three types of images preserve semantic similarities with the original images better, enabling more effective visual contrast through symmetric comparison. In contrast, black images suffer from severe information deficiency, while cropped images risk eliminating key visual elements relevant to the prompt, both factors weakening their effects in visual understanding enhancement. %Overall, the above experiment results suggest using similar, noisy, or synthetic images for multimodal preference alignment yields comparable performance in hallucination mitigation for MLLMs.

\section{Limitation}\label{section:limitation}
Despite its promising performance in mitigating hallucination of MLLMs, SymMPO has two limitations. (1) Due to constraints in our cost-effective and efficient preference data construction pipeline, SymMPO's performance on tasks that involve fine-grained visual understanding (e.g., detailed image description generation) remains limited (discussed in Section~\ref{mainresults}). 
(2) SymMPO introduces additional computational overhead as it requires the construction of preferred response for each contrastive image. Although we have designed a cost-effective caption-anchored response preference pair construction pipeline for reducing the computational cost, we acknowledge it as a limitation. % This pipeline leverage open-source MLLMs alongside efficient and powerful LLMs, such as Qwen2.5-VL and DeepSeek-V3,
%Notably, to address this challenge, we designed a cost-effective caption-anchored response preference pair construction pipeline. This pipeline leverage open-source MLLMs alongside efficient and powerful LLMs, such as Qwen2.5-VL and DeepSeek-V3, to significantly reduce the associated computational burden. % which replace the need for more expensive API calls and extensive multi-stage processing used in previous studies. %Finally, although the two Bradley-Terry models achieve effective visual contrast, the approach of comparing $y_w$ being preferred over $y_w'$ for a given visual-text input and observing preference reversals under contrastive images remains indirect. Further exploration is needed to develop more direct and effective methods for visual contrast in MLLMs.

\section{Conclusion}
In this paper, we propose Symmetric Multimodal Preference Optimization (SymMPO), which effectively addresses the limitations of non-rigorous objective function and indirect preference supervision in existing vision-enhanced multimodal preference optimization methods.  The key novelty lies in the symmetric pair-wise preference optimization formulation and preference margin consistency regularization. 
%Specifically, SymMPO introduces a symmetric pair-wise preference optimization formulation, which effectively utilize the pair-wise preferred responses of contrastive images for enhancing the vision-oriented preference learning.   with Bradley-Terry model for pairwise preference optimization, 
Extensive experiments conducted across five multimodal benchmarks validate the superiority of our SymMPO framework, while ablation studies confirm the necessity of each key component within the framework. Additional comparisons on contrastive image types highlight the robustness of our model with diverse types.% of contrastive images and the advantage of using CLIP to select contrastive images for multimodal preference alignment. %visual understanding enhancement. %, demonstrating superior performance compared to other common approaches.

%%%%%%%%%%%%%%%%%%%%%%%%%%%%%%%%%%%%%%%%%%%%%%%%%%%%%%%%%%%%
\begin{ack}
This research is supported by the National Natural Science Foundation of China (Grants No. 62376137 and 62172261) and the Shandong Provincial Natural Science Foundation (Grant No. ZR2022YQ59), for which we express our heartfelt gratitude. We also deeply appreciate the support and assistance provided by the iLearn Laboratory at Shandong University during the course of this research.

% Use unnumbered first level headings for the acknowledgments. All acknowledgments

% Do {\bf not} include this section in the anonymized submission, only in the final paper. You can use the \texttt{ack} environment provided in the style file to automatically hide this section in the anonymized submission.
\end{ack}

% \section*{References}

% References follow the acknowledgments in the camera-ready paper. Use unnumbered first-level heading for
% the references. Any choice of citation style is acceptable as long as you are
% consistent. It is permissible to reduce the font size to \verb+small+ (9 point)
% when listing the references.
% Note that the Reference section does not count towards the page limit.
% \medskip

{
% \small
\bibliographystyle{unsrt}
\bibliography{neurips_2025}
}

%%%%%%%%%%%%%%%%%%%%%%%%%%%%%%%%%%%%%%%%%%%%%%%%%%%%%%%%%%%%
\newpage
\appendix

\section{Related Work}
Existing approaches that apply DPO to align MLLMs fundamentally involves two components: (1) preference data construction and (2) optimization strategy. We accordingly present current multimodal DPO research from these two perspectives.

%We will subsequently discuss each category in detail.

\subsection{Preference Data Construction}\label{related:preference_data}
%To enhance preference optimization for MLLMs, researchers have proposed various methodologies for generating high-quality preference data. These 
Exiting preference data construction approaches primarily fall into two categories: (1) comparative ranking-based methods and (2) hallucination correction-based methods. %and (3) atomic claim decomposition-based rewriting methods.

 %introduces human annotators into the correction pipeline, employing 
The first category constructs preference data by directly evaluating the rankings of candidate responses. For example, CLIP-DPO~\cite{ouali2024clip} uses CLIP's image-text similarity metrics to rank candidate responses and selects pairs with large ranking gaps and similar lengths as preference data. Beyond this, AMP~\cite{zhang2024automated} leverages MLLMs of varying scales to generate multi-level preference data, assuming that responses from larger MLLMs should rank higher. Furthermore, BPO~\cite{pi2024strengthening} distorts the input image and pairs the response generated from the distorted image (ranked lower) with the response from the original image (ranked higher) to create preference pairs. To eliminate confounding factors such as text style that hinder the model's ability to discern genuine trustworthiness differences within response pairs, RLAIF-V~\cite{yu2024rlaif} introduces a deconfounded candidate response generation strategy. This strategy generates candidate responses through multiple sampling decoding trials with different random seeds while keeping the input prompt and decoding parameters constant. % Additionally, RLAIF-V employs a divide-and-conquer approach for response quality evaluation, which simplifies the complex response-evaluation task by breaking it down into multiple simpler claim-evaluation~tasks.

The second category constructs preference data by detecting and correcting hallucinated content in MLLM responses. For example, RLHF-V~\cite{yu2024rlhf} relies on human annotators to identify and rectify hallucinatory content, ensuring the creation of higher-quality preference datasets. In contrast, HA-DPO~\cite{zhao2023beyond} bypasses human annotators by leveraging GPT-4~\cite{achiam2023gpt} to detect and correct hallucination using fine-grained visual annotations from the Visual Genome dataset~\cite{krishna2017visual}. Similarly, OPA-DPO~\cite{yang2025mitigating} employs GPT-4V~\cite{openai2023gpt4v} to rectify hallucination but identifies them through direct image pair comparisons, eliminating the need for fine-grained annotations and offering greater flexibility. To reduce the high API costs associated with commercial LLMs, HSA-DPO~\cite{xiao2024detecting} trains a specialized hallucination detection model using a sentence-level annotation dataset generated by GPT. %An open-source MLLM then corrects the hallucination to construct the final preference data. 
Moreover, building on RLAIF-V~\cite{yu2024rlaif}, TPO~\cite{he2024topic} introduces a topic-level self-correction paradigm. This method works on identifying hallucination at the topic level through sub-sentence clustering, constructing topic-level preference pairs, and generating response preference pairs using a deconfounded topic-overwriting strategy---ensuring linguistic style consistency.

\subsection{Optimization Strategies}\label{related:optimization_design}
%Building upon recent advances in preference data construction, researchers have significantly improved the preference optimization process for MLLMs by designing various optimization approaches, leading to enhanced optimization outcomes.

Apart from enhancing preference data construction, researchers have also explored various optimization strategies to improve MLLM preference alignment. For example, MPO~\cite{wang2024enhancing} introduces a mixed preference optimization model, which effectively combines preference optimization techniques and conventional supervised fine-tuning. %Specifically, it jointly conducts relative preference learning between pairs of responses, absolute quality evaluation of individual responses, and preferred response generation with the DPO loss, BCO~\cite{jung2024binary} (Binary Classifier Optimization) loss, and supervised fine-tuning loss, respectively.
To mitigate the hallucination of MLLMs, AMP~\cite{zhang2024automated} introduces a multi-level direct preference optimization algorithm, enabling robust multi-level preference learning, while CHiP~\cite{fu2025chip} introduces a cross-modal hierarchical DPO model involving two key optimizations: hierarchical textual preference optimization for capturing fine-grained textual preferences and visual preference optimization for cross-modal preference alignment. 
To address the gradient vanishing problem induced by off-policy data, OPA-DPO~\cite{yang2025mitigating} proposes an adaptive mechanism that dynamically balances exploration and exploitation during learning. 
Furthermore, MIA-DPO~\cite{liu2024mia} specifically targets hallucination reduction in multi-image scenarios, where MLLMs process multiple input images simultaneously through optimized cross-image attention mechanisms.

One key issue in improving MLLMs' reasoning ability is mitigating their over-reliance on textual prompts while enhancing visual content utilization. To address this issue, SymDPO~\cite{jia2024symdpo} introduces a symbol demonstration direct preference optimization model for in-context learning, which strengthens MLLMs' visual understanding by replacing textual answers with random symbols, thereby forcing MLLMs to establish mappings between visual information and symbolic responses. %Notably, this approach adopts the conventional DPO loss and only supports in-context learning.
For more general multimodal understanding contexts, several studies~\cite{wang2024mdpo, xie2024v, jiang2024modality, fu2025chip, yang2025mitigating} focused on extending DPO with vision-oriented preference learning to improve MLLMs' visual signal interpretation. These approaches preserve DPO's structural framework while only introducing visual variation in the contrastive image-prompt-response triplet pairs. % where each data pair shares a common input textual prompt and target response but differs in the input visual content.
For contrastive image generation, existing methods employ diverse strategies: mDPO~\cite{wang2024mdpo} applies geometric transformations to original images, V-DPO~\cite{xie2024v} employs a generative model to replace key visual elements through image inpainting~\cite{lugmayr2022repaint},  MFPO~\cite{jiang2024modality}, OPA-DPO~\cite{yang2025mitigating} and CHiP~\cite{fu2025chip} perform noise injection to original images, while CHiP~\cite{fu2025chip} utilizes a forward diffusion process~\cite{ho2020denoising} for generating contrastive images. %, progressively adding Gaussian noise to the given image. %By combining these vision-oriented preference learning with standard DPO optimization, these methods achieve simultaneous learning from both textual and visual preference signals, substantially enhancing response quality and multimodal comprehension.

Although current vision-enhanced multimodal preference alignment methods have demonstrated great progress in reducing MLLM hallucination, they exhibit two critical limitations: (1) their DPO-based objective function derivations lack mathematical rigor, as they overlook the fact that the intractable partition functions in multimodal scenarios with different vision inputs cannot be directly canceled; 
and (2) they fail to provide direct preference supervision for DPO-based visual understanding enhancement. % which deviates from  DPO's core principle of utilizing response pairs as preference.
%reward formulation by failing to explicitly model the direct preference relationship between accurate and hallucinated response. 
To address these limitations, we propose Symmetric Multimodal Preference Optimization, which effectively utilize the corresponding preferred responses of contrastive images for optimizing the visual understanding capabilities of MLLMs. 

%jointly mines contrastive response preference relationships across two image-prompt-response triplets involving both different  sharing the same prompt.
 
%These limitations underscore the necessity for more principled training objectives that properly account for the unique characteristics of multimodal learning and the complex relationships between visual and textual modalities.

\section{Impact of Partition Function in Multimodal Preference Optimization}\label{appendix:discussion_of_partition_function}
Due to space limitations, Section~\ref{section:introduction} briefly argues that existing multimodal DPO methods~\cite{wang2024mdpo, xie2024v, jiang2024modality, fu2025chip, yang2025mitigating} non-rigorously ignore two partition functions in their vision-oriented contrastive learning mechanisms. Here, we present a detailed analysis of these functions' roles in model optimization.

Specifically, according to the standard DPO formulation (Equation~\ref{Formula:implicit_reward}), the implicit reward of MLLMs can be expressed as follows:
\begin{equation}
\left\{\begin{aligned}
    &r(m,x,y)=\beta\log\frac{\pi_\theta(y\vert m,x)}{\pi_{ref}(y\vert m,x)}+\beta\log Z(m,x),\\
     &Z(m,x)=\sum_y\pi_{ref}(y\vert m,x)\exp\big(\frac{1}{\beta}r(m,x,y)\big),
\end{aligned}\right.
\label{Formula:implicit_reward_of_MLLMs}
\end{equation}
where  $m$, $x$, and $y$ denote the input image, textual prompt, and corresponding response, respectively. $\pi_\theta$, $\pi_{ref}$, and $r$ are the policy model, reference model, and implicit reward function, respectively. $Z(m,x)$ is the partition function for the multimodal scenario, derived by incorporating an image variable into the single-modal partition function defined by standard DPO~\cite{rafailov2023direct}.

%substitute the implicit reward formulation into the Bradley-Terry model (Equation~\ref{Formula:Bradley_Terry_model})  to derive the corresponding 
Adopting the mainstream vision-oriented preference learning  paradigm that introduces additional contrastive images for preference alignment, we derive the following 
loss function by substituting the above implicit reward formulation into the Bradley-Terry model (Equation~\ref{Formula:Bradley_Terry_model}),
%P_{BT}(y_w\succ y_w'\vert m,x)\wedge P_{BT}(y_w'\succ y_w\vert m',x) \\    =\ \cdot\sigma\big(r(m',x,y_w')-r(m',x,y_w)\big).
\begin{equation}
\begin{aligned}
    \mathcal{L}_{VCO}^*=&-\mathbb{E}_{(x,m_w,m_l,y_w)\sim\mathcal{D}}\Big[\log\sigma\big(r(m_w,x,y_w)-r(m_l,x,y_w)\big)\Big]\\
    =&-\mathbb{E}_{(x,m_w,m_l,y_w)\sim\mathcal{D}}\Big[\log\sigma\Big(\beta\log\frac{\pi_\theta(y_w\vert m_w,x)}{\pi_{ref}(y_w\vert m_w,x)}-\beta\log\frac{\pi_\theta(y_w\vert m_l,x)}{\pi_{ref}(y_w\vert m_l,x)}\\
    &\hspace{3cm}{+\beta\log Z(m_w,x)-\beta\log Z(m_l,x)}\Big)\Big].
\end{aligned}
\label{Formula:vco_with_partition_function}
\end{equation}
Existing methods directly cancel out $Z(m_w,x)$ and $Z(m_l,x)$ in Equation~\ref{Formular:vision}, which is apparently inappropriate based on the above rigorous derivation. 

To better understand the role of partition functions in the preference learning process, we calculate the gradient of $\mathcal{L}_{VCO}^*$ with respect to $\theta$. To facilitate the gradient calculation, for each data instance sampled during training, $(x,m_w,m_l,y_w)\sim\mathcal{D}$, we first define: 
\begin{equation}
\left\{
\begin{aligned}
    u&=\beta\log\frac{\pi_\theta(y_w\vert m_w,x)}{\pi_{ref}(y_w\vert m_w,x)}-\beta\log\frac{\pi_\theta(y_w\vert m_l,x)}{\pi_{ref}(y_w\vert m_l,x)},\\
    c&=\beta\log Z(m_w,x)-\beta\log Z(m_l,x).  
\end{aligned}\right.
\nonumber
\end{equation}
where $u$ involves the policy model optimization, while $c$ does not. In fact, $c$ remains constant across different policy model parameters $\theta$ because $Z(m,x)$ is independent of $\theta$ in its calculation.
%both the reference model and reward function are fixed during policy model optimization, and  $Z(m,x)$ invariant with respect to $\theta$. 
%The $\pi_{ref}$ represents the reference model, and $r$ denotes the reward function, both of which remain fixed during training. Therefore, for any image $m$ and textual prompt $x$, their partition function remains constant throughout training.

%Since $Z(m,x)$ mathematically integrates over all possible responses $y$, it inherently serves as the normalization constant for the entire response quality distribution conditioned on $(m,x)$. 
Accordingly, the gradient of $\mathcal{L}_{VCO}^*$  with respect to $\theta$ is then given by:
\begin{equation}
\begin{aligned}
    \frac{\partial\mathcal{L}^*_{VCO}}{\partial\theta}&=\frac{\partial\mathcal{L}^*_{VCO}}{\partial u}\cdot\frac{\partial u}{\partial\theta} \\
    &=-\frac{\partial}{\partial u}\log\sigma(u+c)\cdot\frac{\partial u}{\partial\theta} \\
    &=-\big(1-\sigma(u+c)\big)\cdot\frac{\partial u}{\partial\theta} \\
    &=-\sigma\big(-(u+c)\big)\cdot\frac{\partial u}{\partial\theta}.
\end{aligned}
\label{formula:gradient_of_VCO^*}
\end{equation}

Similarly, we can derive the gradient of the visual-oriented contrastive objective used in existing works (i.e., Equation~\ref{Formular:vision}) as:
\begin{equation}
\begin{aligned}
    \frac{\partial\mathcal{L}_{VCO}}{\partial\theta}=-\sigma(-u)\frac{\partial u}{\partial \theta}.
\end{aligned}
\label{formula:gradient_of_VCO}
\end{equation}

By comparing the gradients in Equations~\ref{formula:gradient_of_VCO^*} and~\ref{formula:gradient_of_VCO}, we observe that the constant $c$ in Equations~\ref{formula:gradient_of_VCO^*} acts as an offset modulating the coefficient of the gradient term $\frac{\partial u}{\partial\theta}$. Intuitively, since $Z(m,x)$ integrates over all possible responses $y$, it inherently captures the global quality of the response space conditioned on $(m,x)$. Thus,  $c$ reflects the reference model's global response quality discrepancy 
between the contexts $(m_w, x)$ and $(m_l, x)$. Specifically, a larger value of $c$ indicates a stronger discrepancy in model behavior between $(m_w, x)$ and $(m_l, x)$, exerting greater influence on gradient adjustment for preference alignment optimization.
Conversely, a smaller value of $c$ implies a weaker response quality gap between the multimodal context pair, less impacting the gradient update during training. However, existing methods neglect the two partition functions in visual-oriented contrastive optimization. In essence, these methods assign the same value (i.e., $c=0$) to all contrastive samples, preventing adaptive weight adjustment for different image contexts. Consequently, the model fails to achieve optimal visual understanding capability, as it tends to either over-attend to images with limited informative signals or under-reason about images containing complex visual clues. %" As a result, the model may over-attend to images with limited informative signals, or under-reason about images containing complex visual clues.

\section{Additional Ablation Studies}

\begin{table}
  \caption{Comparison of different learning rates on the performance of SymMPO}
  \label{table:additional_ablation_1}
  \centering
  \begin{adjustbox}{width=\textwidth}
      \begin{tabular}{l*{10}{c}}
        \toprule
        \multirow{2}{*}{SymMPO} & \multicolumn{3}{c}{HallusionBench} & \multicolumn{2}{c}{Object-HalBench} & \multicolumn{2}{c}{MMHal-Bench} & \multicolumn{2}{c}{AMBER} & MMStar \\ \cmidrule(r){2-4} \cmidrule(r){5-6} \cmidrule(r){7-8} \cmidrule(r){9-10} \cmidrule(r){11-11}
        & qAcc$\uparrow$ &fAcc$\uparrow$ &aAcc$\uparrow$ & Resp.$\downarrow$ & Ment.$\downarrow$ & Score$\uparrow$ & Hall$\downarrow$ & Acc$\uparrow$ & F1$\uparrow$ & Overall$\uparrow$ \\
        \midrule
        lr=5e-5 & \textbf{8.13} & 10.40 & 41.09 & 22.6 & 13.9 & 2.29 & 54.2 & \underline{80.8} & 86.1 & 28.2 \\
        lr=5e-6 & \underline{7.25} & \textbf{13.58} & \textbf{44.28} & \textbf{19.5} & \textbf{9.7} & \textbf{2.89} & \textbf{42.7} & \textbf{82.6} & \textbf{87.7} & \textbf{34.8} \\
        lr=5e-7 & \textbf{8.13} & \underline{12.42} & \underline{43.75} & \underline{20.1} & \underline{9.8} & \underline{2.80} & \underline{49.0} & \underline{80.8} & \underline{86.8} & \underline{33.8} \\
        \bottomrule
      \end{tabular}
  \end{adjustbox}
\end{table}

\begin{table}
  \caption{Comparison of different $\lambda$ on the performance of SymMPO}
  \label{table:additional_ablation_2}
  \centering
  \begin{adjustbox}{width=\textwidth}
      \begin{tabular}{l*{10}{c}}
        \toprule
        \multirow{2}{*}{SymMPO} & \multicolumn{3}{c}{HallusionBench} & \multicolumn{2}{c}{Object-HalBench} & \multicolumn{2}{c}{MMHal-Bench} & \multicolumn{2}{c}{AMBER} & MMStar \\ \cmidrule(r){2-4} \cmidrule(r){5-6} \cmidrule(r){7-8} \cmidrule(r){9-10} \cmidrule(r){11-11}
        & qAcc$\uparrow$ &fAcc$\uparrow$ &aAcc$\uparrow$ & Resp.$\downarrow$ & Ment.$\downarrow$ & Score$\uparrow$ & Hall$\downarrow$ & Acc$\uparrow$ & F1$\uparrow$ & Overall$\uparrow$ \\
        \midrule
        $\lambda$=0.1 & 5.27 & 10.98 & 41.27 & 21.2 & 12.0 & 2.61 & 50.0 & 80.6 & 86.4 & 34.2 \\
        $\lambda$=0.3 & \textbf{7.47} & 10.98 & \underline{42.60} & 19.7 & 10.5 & \textbf{3.07} & \textbf{37.5} & 81.5 & 86.7 & \underline{34.6} \\
        $\lambda$=0.5 & \underline{7.25} & \textbf{13.58} & \textbf{44.28} & \underline{19.5} & \underline{9.7} & \underline{2.89} & \underline{42.7} & \textbf{82.6} & \textbf{87.7} & \textbf{34.8} \\
        $\lambda$=0.7 & 5.93 & \underline{12.71} & 41.36 & \textbf{18.1} & \textbf{9.3} & 2.83 & 44.8 & \textbf{82.6} & \textbf{87.7} & 34.4 \\
        $\lambda$=0.9 & \textbf{7.47} & 10.98 & 41.98 & 21.2 & 10.7 & 2.76 & 46.9 & \underline{81.8} & \underline{87.2} & 34.2 \\
        \bottomrule
      \end{tabular}
  \end{adjustbox}
\end{table}

\begin{table}
  \caption{Comparison of different $\gamma$ on the performance of SymMPO}
  \label{table:additional_ablation_3}
  \centering
  \begin{adjustbox}{width=\textwidth}
      \begin{tabular}{l*{10}{c}}
        \toprule
        \multirow{2}{*}{SymMPO} & \multicolumn{3}{c}{HallusionBench} & \multicolumn{2}{c}{Object-HalBench} & \multicolumn{2}{c}{MMHal-Bench} & \multicolumn{2}{c}{AMBER} & MMStar \\ \cmidrule(r){2-4} \cmidrule(r){5-6} \cmidrule(r){7-8} \cmidrule(r){9-10} \cmidrule(r){11-11}
        & qAcc$\uparrow$ &fAcc$\uparrow$ &aAcc$\uparrow$ & Resp.$\downarrow$ & Ment.$\downarrow$ & Score$\uparrow$ & Hall$\downarrow$ & Acc$\uparrow$ & F1$\uparrow$ & Overall$\uparrow$ \\
        \midrule
        $\gamma$=1e-2 & 4.39 & 10.40 & 40.83 & 24.4 & 15.2 & 2.15 & 58.3 & 78.9 & 83.5 & \textbf{34.9} \\
        $\gamma$=1e-3 & \underline{6.59} & 10.40 & 43.13 & \underline{21.0} & \underline{9.9} & \underline{2.70} & 49.0 & \textbf{83.1} & 87.6 & \textbf{34.9} \\
        $\gamma$=1e-4 & \textbf{7.25} & \textbf{13.58} & \textbf{44.28} & \textbf{19.5} & \textbf{9.7} & \textbf{2.89} & \textbf{42.7} & \underline{82.6} & \underline{87.7} & \underline{34.8} \\
        $\gamma$=1e-5 & 6.37 & \underline{10.98} & \underline{43.75} & \underline{21.0} & 10.4 & \underline{2.70} & \underline{47.9} & 82.4 & \textbf{87.8} & 33.2 \\
        \bottomrule
      \end{tabular}
  \end{adjustbox}
\end{table}

In this section, we detailed the process of selecting hyper-parameters for our experiments.

For the hyper-parameters $\beta$, $\eta$ and $\delta$, we adopt the same setup as prior works. Specifically, $\beta$ is directly set to 0.1, while $\eta$ and $\delta$ are chosen to be 1.0 and 0, respectively, in alignment with the configurations used in mDPO and OPA-DPO. For the other hyper-parameters, including the learning rate, $\lambda$ and $\gamma$, we determine their values through a grid search. The search is conducted over the following ranges: the learning rate is evaluated at [5e-5, 5e-6, 5e-7], $\lambda$ is tested across [0.1, 0.3, 0.5, 0.7, 0.9], and $\gamma$ is explored within [1e-2, 1e-3, 1e-4, 1e-5]. The results from this grid search are presented in Table~\ref{table:additional_ablation_1}, Table~\ref{table:additional_ablation_2}, and Table~\ref{table:additional_ablation_3}, respectively, with all experiments conducted using LLaVA-1.5-7B. Based on these results, the final hyper-parameter values chosen for our experiments are as follows: a learning rate of 5e-6, $\lambda$=0.5, and $\gamma$=1e-4.

Compared to the learning rate and $\lambda$, the hyper-parameter $\gamma$, which governs preference margin consistency regularization, is notably smaller. This may be attributed to the subtle and sensitive nature of relative relationships between preferences across contrastive images. As a result, smaller $\gamma$ values lead to weaker regularization of preference margin consistency, limiting the effectiveness of SymMPO. Conversely, larger $\gamma$ values impose excessively strong regularization, which can disrupt preference learning. These observations highlight the critical need to carefully balance the regularization strength to achieve optimal performance.

\section{Case Study}
In this part, we present the case study on our constructed preference pair data for model training and generated responses for model evaluation. 

\begin{figure}
    \centering
    \includegraphics[width=0.95\linewidth]{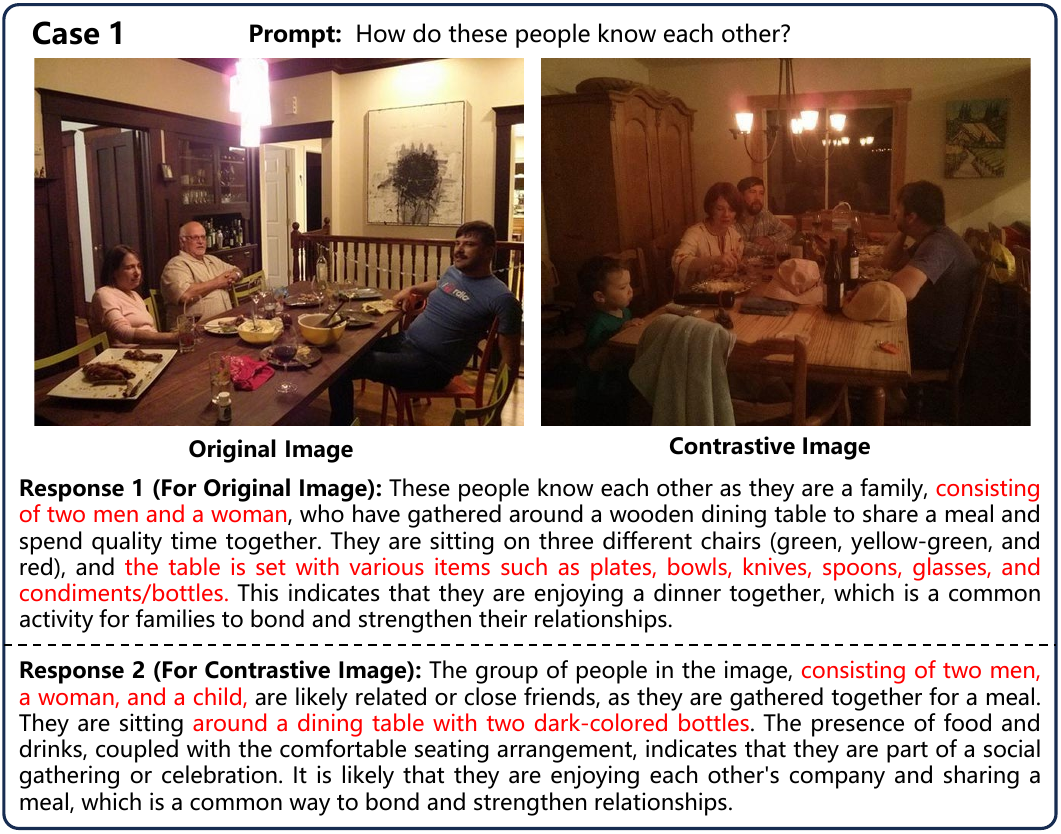}
    \hfill
    \includegraphics[width=0.95\linewidth]{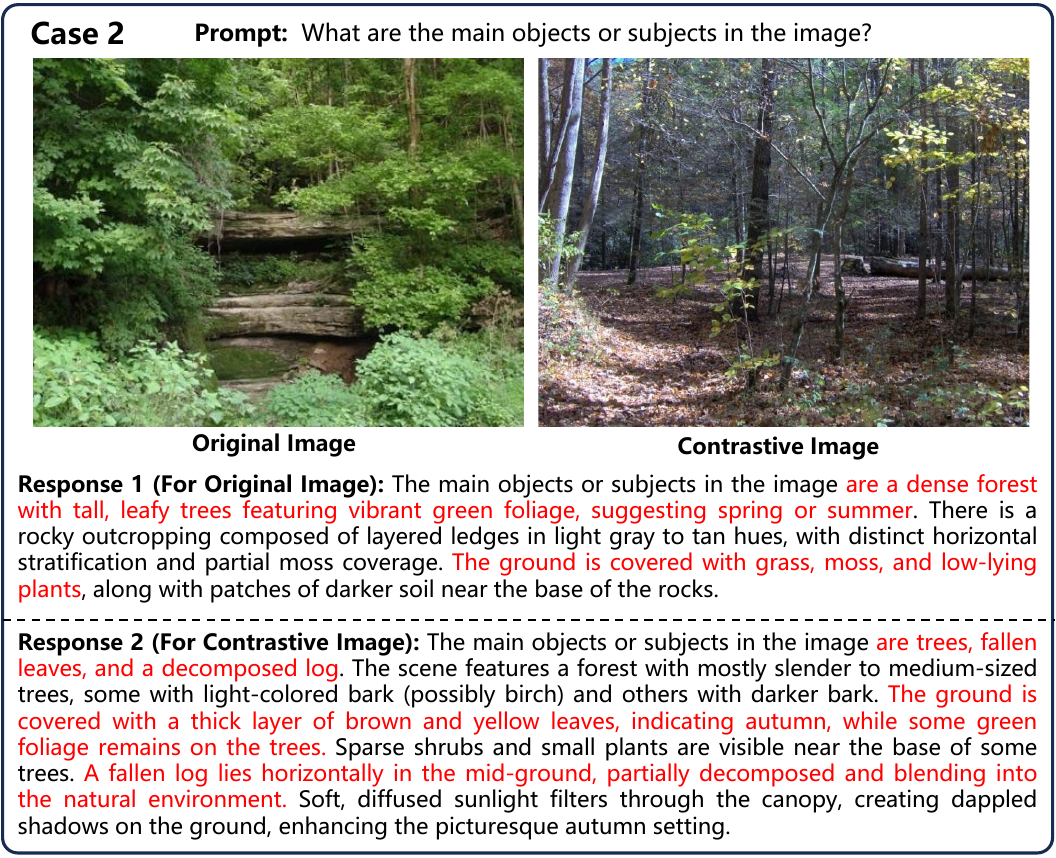}
    \caption{Two training samples from our built preference dataset used to optimize SymMPO, with hallucinated elements relative to the other response highlighted in red.}
    \label{figure:data_sample}
\end{figure}

%we select two training data samples from the dataset, constructed by pairing original images with the most contrastive images identified using CLIP similarity scores across the entire dataset, along with their corresponding responses generated using the data construction pipeline described in Section~\ref{data_construction}. 
\textbf{On Preference Pair Construction.} As illustrated in Figure~\ref{fig:symmpo}, each training sample for our symmetric preference optimization comprises two triplets sharing the same prompt and similar but distinct image-response pairs. Empirically, the contrastive images are identified using CLIP similarity scores, while the responses are generated using the caption-anchored response preference pair construction pipeline described in Section~\ref{data_construction} for corresponding images given the same prompt. Figure~\ref{figure:data_sample} illustrates two training examples from our constructed preference dataset, where claims in each response that contradict (and are thus potentially hallucinated relative to) the paired response are highlighted in red. These two examples demonstrate that the effectiveness of using CLIP similarity to obtain contrastive image pairs with subtle visual differences, and confirm that the corresponding responses generated by our preference data construction pipeline indeed exhibit strong linguistic alignment with only minor claim variations. Overall, these examples show the effectiveness of our training preference pair construction strategy.  %Combining experiment results in Section 4.3, we can conclude that training models with such triplet pairs containing similar contrastive images and responses can improve the model's visual understanding capabilities and alleviate the hallucination issue. 
%vision understanding capabilities learning can be effectively improved and the hallucination issue gets alleviated.
%By comparing such minor claim variations between responses during the preference learning, the model's visual understanding capabilities can be effectively improved and the hallucination issue gets alleviated. 

%The subtle variations in an response can be indeed interpreted as the potential hallucinated content of MLLMs for generating response for the other image-prompt input. Therefore, utilizing response pairs that contain mutually hallucinated content for model optimization can strengthen the MLLM's preference alignment and mitigate its hallucination.

%To gain an intuitive understanding of the practical preference data we built for model optimization, we show two training preference pair examples in Figure~\ref{figure:data_sample}, where the potentially hallucinated claims in each response are marked in green.

\begin{figure}
    \centering
    \includegraphics[width=1.0\linewidth]{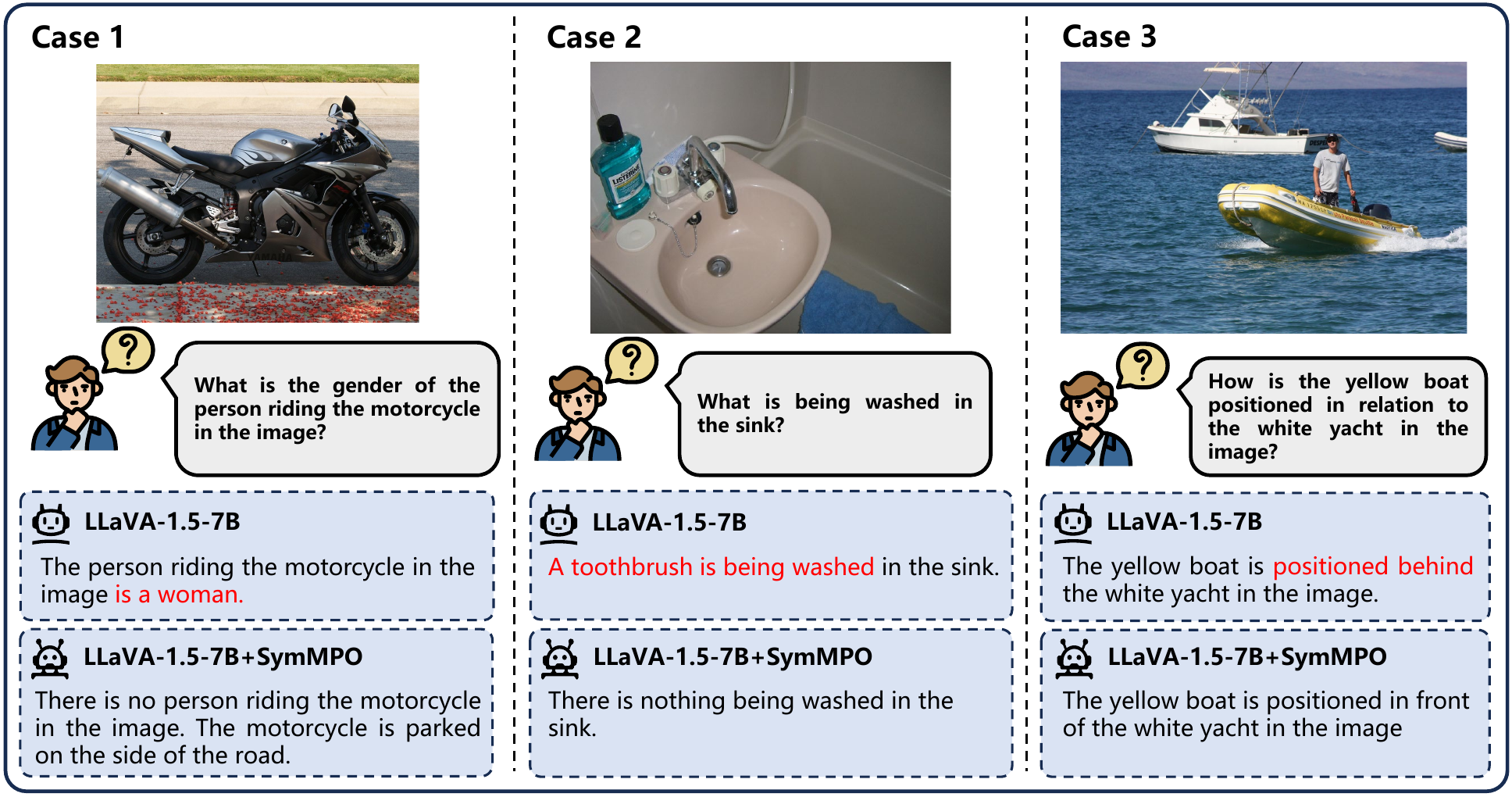}
    \caption{Responses generated by LLaVA-1.5-7B and our SymMPO-enhanced version for data examples from MMHal-Bench, with hallucinated contents highlighted in red.}
    \label{figure:case_study_mmhal} %\vspace{-0.8em}
\end{figure}

\begin{figure}
    \centering
    \includegraphics[width=1.0\linewidth]{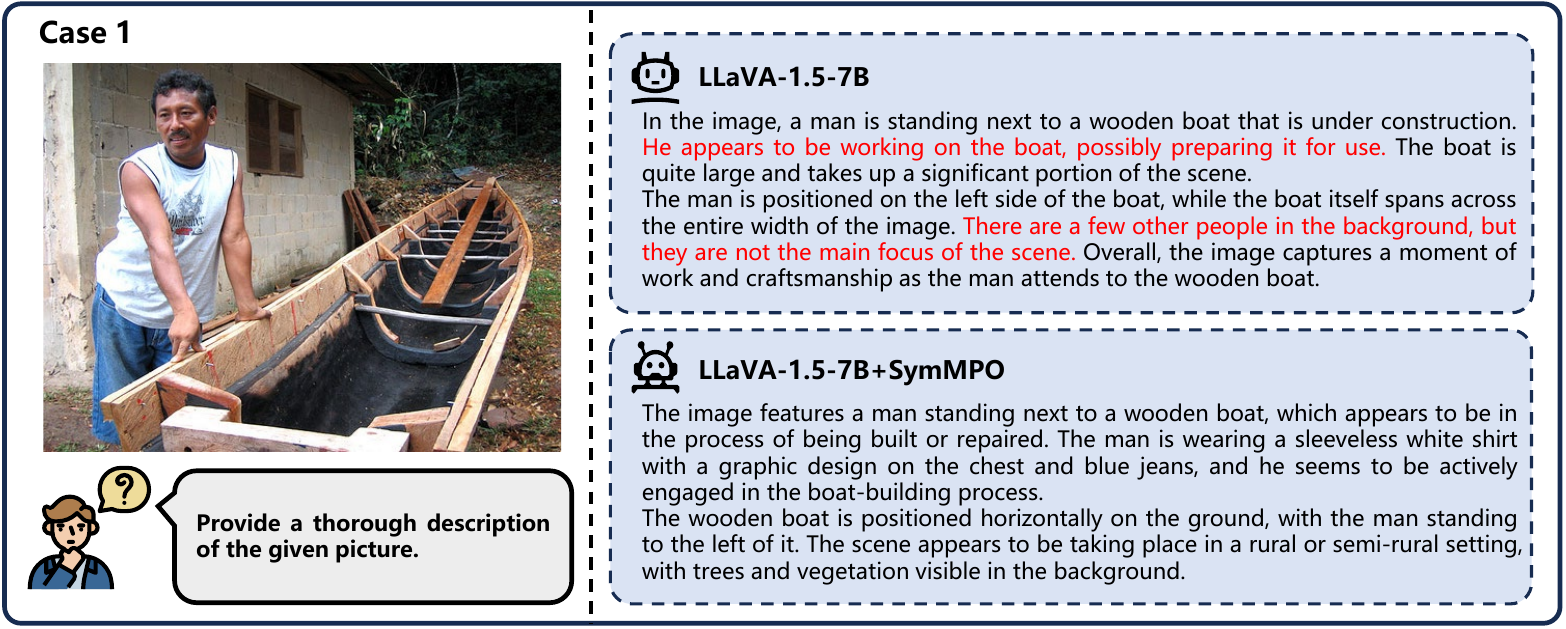}
    \hfill
    \includegraphics[width=1.0\linewidth]{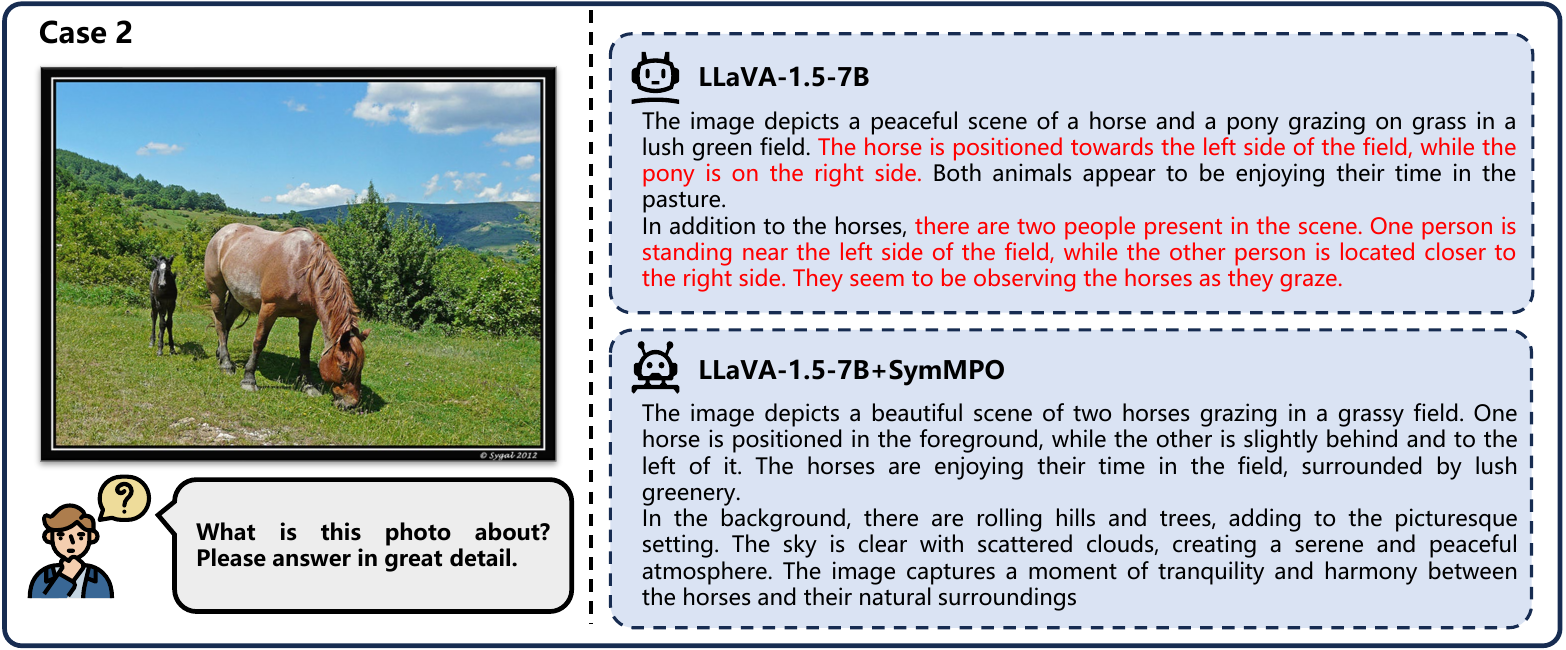}
    \caption{Responses generated by LLaVA-1.5-7B and our SymMPO-enhanced version for data examples from Object-HalBench, with hallucinated contents highlighted in red.}
    \label{figure:case_study_objhal_1}
\end{figure}

\textbf{On Hallucination Mitigation.} %Using these benchmarks, responses were generated by LLaVA-1.5-7B and SymMPO. The examples from MMHal-Bench are shown in Figure~\ref{figure:case_study_mmhal}, while the examples from Object-HalBench are illustrated in 
To demonstrate the advantages of SymMPO, we compare the specific responses generated by original LLaVA-1.5-7B and our SymMPO-enhanced version with samples from two widely used benchmarks, MMHal-Bench and Object-HalBench. Both benchmarks feature diverse visual questions closely related to daily life. 
Figure~\ref{figure:case_study_mmhal} and Figure~\ref{figure:case_study_objhal_1} present the corresponding results, with hallucinated content highlighted in red.  
%Both benchmarks features visual questions closely related to daily life. Figure~\ref{figure:case_study_mmhal} and Figure~\ref{figure:case_study_objhal_1} show the response generation results on MMHal-Bench and Object-HalBench, respectively, with hallucinated content is highlighted in red.
As can be seen,  LLaVA-1.5-7B+SymMPO 
consistently generates more accurate responses than the original  LLaVA-1.5-7B across both benchmarks. Even for visual questions  requiring detailed responses in Object-HalBench, LLaVA-1.5-7B+SymMPO maintains precise, hallucination-free responses, while LLaVA-1.5-7B fails by producing responses with hallucinations.
These results confirm the effectiveness of our SymMPO in enhancing vision-language understanding and mitigating MLLM hallucinations.

\end{document}